%% file: acl_latex.tex
\documentclass[11pt]{article}

\usepackage[preprint]{acl}

\usepackage{times}
\usepackage{latexsym}

\usepackage[T1]{fontenc}

\usepackage[utf8]{inputenc}

\usepackage{microtype}

\usepackage{inconsolata}

\PassOptionsToPackage{hyphens}{url}
\usepackage{hyperref}

\usepackage{graphicx}
\usepackage{booktabs}
\usepackage{float}

\usepackage{multicol}
\usepackage{multirow}
\usepackage{amsmath,mathtools}
\usepackage{subcaption}

\usepackage{acro}
\DeclareAcronym{IBN}{short=IBN,long=Intent Based Networking}
\DeclareAcronym{NLP}{short=NLP,long=Natural Language Processing}
\DeclareAcronym{LLM}{short=LLM,long=Large Language Model}
\DeclareAcronym{BERT}{short=BERT,long=Bidirectional Encoder Representations from Transformers}
\DeclareAcronym{GPT}{short=GPT,long=Generative Pre-trained Transformer}
\DeclareAcronym{Nile}{short=Nile,long=Network Intent Language}
\DeclareAcronym{QoS}{short=QoS, long=Quality of Service}
\DeclareAcronym{ACL}{short=ACL, long=Access Control List}
\DeclareAcronym{RAG}{short=RAG, long=Retrieval Augmented Generation}
\DeclareAcronym{NEAT}{short=NEAT, long=Nile-English Aligned Translations}
\DeclareAcronym{llmjudge}{short=LLM-judge, long=LLM-as-a-judge}
\DeclareAcronym{SOTA}{short=SOTA, long=state-of-the-art}

\usepackage{listings}
\usepackage[most]{tcolorbox}
\newtcblisting[auto counter]{codelisting}[2][]{sharp corners, 
    fonttitle=\footnotesize\bfseries, colframe=gray, listing only, 
    listing options={basicstyle=\footnotesize\ttfamily, breaklines=true, breakautoindent=false, breakindent=0pt, escapeinside={(*@}{@*)}
    },
    title=Prompt segment \thetcbcounter: #2, #1}

\lstnewenvironment{pycode}{
  \lstset{
    language=Python,
    basicstyle=\scriptsize\ttfamily,
    breaklines=true,
    escapeinside={(*@}{@*)},
    commentstyle=\color{green!60!black}\itshape,
    keywordstyle=\color{gray},
    stringstyle=\color{black},
    showstringspaces=false,
    emph={scorer, S, H, score, template, device},
    emphstyle=\color{blue!70!black},
    emph={[2]ParaPLUIE, init, setTemplate, compute, print},
    emphstyle={[2]\color{orange!80!black}},
  }
}{}

\newcommand{\loss}{\textrm{$loss$}}
\newcommand{\Prompt}{\textrm{Prompt}}
\newcommand{\llm} {\textit{LLM}}
\newcommand{\llmlikert}{LLM-Likert}
\newcommand{\llmchoice}{LLM-choice}

\newcommand{\PPluie}{ParaPLUIE}
\newcommand{\tplPlui}[1]{#1-PLUIE}
\newcommand{\ModBScore}{Modern BertScore}
\newcommand{\Bleu}{BLEU}
\newcommand{\BertScore}{BERTScore}
\newcommand{\Lev}{Levenshtein}
\newcommand{\Meteor}{METEOR}

\newcommand{\phiquatre}{Phi}
\newcommand{\LlamaH}{Llama 3 8B}
\newcommand{\LlamaS}{Llama}
\newcommand{\mistral}{Mistral 7B}
\newcommand{\gptIVmini}{GPT-4o mini}
\newcommand{\gptIV}{GPT-4o}

\newcommand{\phiquatreDef}{Phi-4 14B}
\newcommand{\LlamaSDef}{Llama 3 70B}

\title{*-PLUIE: \underline{P}ersonalisable metric with \underline{L}lm \underline{U}sed for \underline{I}mproved \underline{E}valuation}

\author{
  \textbf{Quentin Lemesle\textsuperscript{1}},
  \textbf{Léane Jourdan\textsuperscript{2}},
  \textbf{Daisy Munson\textsuperscript{3}},
  \textbf{Pierre Alain\textsuperscript{3}},
\\ 
  \textbf{Jonathan Chevelu\textsuperscript{1}},
  \textbf{Arnaud Delhay\textsuperscript{1}},
  \textbf{Damien Lolive\textsuperscript{4}}
\\
\\
  \textsuperscript{1}Univ Rennes, CNRS, IRISA, EXPRESSION, 22300 Lannion, France \\
  \textsuperscript{2}Nantes Université, École Centrale Nantes, CNRS, LS2N, UMR 6004, F-44000 Nantes, France\\
  \textsuperscript{3}Univ Rennes, CNRS, IRISA, SOTERN, 22300 Lannion, France \\
  \textsuperscript{4}Univ of South Brittany, CNRS, IRISA, ARCHIMEDIA, 56000 Vannes, France \\
  \small{
   \textbf{Correspondence:} \href{mailto:quentin.lemesle@irisa.fr}{quentin.lemesle@irisa.fr}
  }
}

\begin{document}
\maketitle
\begin{abstract}

Evaluating the quality of automatically generated text often relies on \ac{llmjudge} methods.
While effective, these approaches are computationally expensive and require post-processing.
To address these limitations, we build upon \PPluie{}, a perplexity-based \ac{llmjudge} metric that estimates confidence over ``Yes/No'' answers without generating text.
We introduce \tplPlui{*}, task-specific prompting variants of \PPluie{} and evaluate their alignment with human judgement.
Our experiments show that personalised \tplPlui{*} achieves stronger correlations with human ratings while maintaining low computational cost.
\end{abstract}

\section{Introduction}

Automatic evaluation is still a challenge in free-form text generation. Traditional similarity-based metrics focus on surface-level lexical overlap and often fail to capture meaning-preserving variations or stylistic improvements.
Recent advances in Large Language Models have introduced \ac{llmjudge} methods, which use the reasoning capabilities of LLMs for interpreting user demands to evaluate generated text~\citep{nips_fslearners}.
They leverage the semantic understanding and contextual reasoning of LLMs to provide richer, more human-aligned assessments across a variety of NLP tasks~\citep{doostmohammadi-etal-2024-reliable}.

However, standard \ac{llmjudge} methods~\cite{gu2025surveyllmasajudge} generate free-form text responses that need to be parsed into structured judgements. 
This parsing introduces noise and ambiguity, especially when simple categorical decisions (e.g. ``Yes/No'') are required. 
This text-generation process may not fully exploit the model’s internal knowledge.
To mitigate these limitations,~\citet{lemesle-etal-2025-paraphrase} proposed \PPluie{} as a perplexity-based alternative to output-based LLM judging, which is formally defined in Appendix \ref{apx:pplui_def}.
It grants a confidence score for a ``Yes/No'' question by relying on the perplexity of the LLM.
Originally introduced for paraphrase classification, \PPluie{} achieves strong alignment with human binary annotations and with minimal computational cost, roughly equivalent to generating one token.
By definition, \PPluie{} represents the model's confidence in its answer, offering several appealing interpretability properties.
Intuitively, a strong positive and negative score should indicate high confidence in “Yes” and “No” answers respectively. 
However, this property has not been evaluated, and it has not been compared to output-based \ac{llmjudge}. 

We also evaluate its applicability to other tasks, language and its alignment to human judgements on preference-based evaluation and compare it to output-based \ac{llmjudge}. 
Specifically, we investigate whether task-specific prompting (\tplPlui{*}), i.e., tailoring the question to the evaluation context, can improve the reliability and generalisation of \PPluie{}.
We conduct this study on three semantic tasks:
French \textbf{Paraphrase classification}, evaluation of \textbf{\ac{Nile} translation} and \textbf{Scientific text revision} quality.
For each task, we compare \tplPlui{*} against widely used similarity-based metrics, output-based \ac{llmjudge} and a random-based approach.
Our main contributions are as follows: 
\textbf{(1)} We introduce \tplPlui{*}, a general and personalisable perplexity-based method for \ac{llmjudge}.
\textbf{(2)} We design and evaluate three task-specific variants of \tplPlui{*}, covering three semantic tasks, to assess the adaptability of the approach.
\textbf{(3)} We show that \tplPlui{*} achieves stronger alignment with human judgement while being up to almost 8 times faster to compute compared to other \ac{llmjudge} metrics.

\section{Experimental Protocol}

We describe the tasks and data considered, the baseline metrics, and discuss how the \PPluie{} methodology is adapted through prompt design.

\subsection{Semantic Tasks}

\textbf{Paraphrase Classification:}
As \PPluie{} demonstrated strong performance in English paraphrase classification~\citep{lemesle-etal-2025-paraphrase}, we examine its generalisation capabilities to French.
For this purpose, we use the dataset proposed by \citet{tytgat-etal-2024-evaluation} which contains French sentences manually transformed with synonym and paronym substitution. From this, we create a French paraphrase dataset that contains 33.60\% positive pairs, and describe the methodology in Appendix~\ref{apx:fr_para_dataset}.

\textbf{Nile Translation:}
\ac{Nile}~\citep{jacobs2018:nile}, provides a structured yet flexible grammar for expressing access control, quality of service, and temporal statements, making it well-suited for bridging the gap between natural language inputs and enforceable network policies.
\citet{MUNSON2025} introduced \ac{NEAT}, a methodology used to create a large-scale corpus of aligned English–\ac{Nile} intents. 
We use their human evaluation of 436 translation triplets of \ac{Nile} expressions.
Following the \ac{NEAT} protocol, Mushra scores of Nile translations are binarised using an acceptability cutoff: translations rated ``Good'' or ``Excellent'' are considered positive, leading to 60\% positive examples. 

\textbf{Scientific Text Revision:}
Text revision is a writing assistance task that involves substantially modifying an existing text to improve it while preserving its original meaning~\citep{du-etal-2022-read,li-etal-2022-text}. 
In the scientific domain, revision is a critical step of the writing process ensuring clarity, coherence, and adherence to academic standards, as poor writing quality can contribute to paper rejection~\citep{amano2023manifold}.
We focus on the paragraph-level scientific text
revision task~\citep{jourdan-etal-2025-pararev}.
We use the test split of the ParaReval  dataset\footnote{\url{https://github.com/JourdanL/parareval}}~\citep{jourdan-etal-2025-identifying}, a collection of human pairwise evaluations of automatically generated revisions. 
More details about this dataset are provided in Appendix~\ref{apx:exp_revision}. 

\begin{table*}[!ht]
  \centering
    \begin{small}
        \input{tables/tytgat_distrib_table}
    \end{small}
  \caption{Natural and \textit{a posteriori} optimal threshold for each metric and model-prompt variant, with corresponding accuracy, precision, recall, and F1-score on classification. 
    \tplPlui{*} is respectively \tplPlui{Fr} and \tplPlui{Net}.
    Confidence interval of reported accuracy, precision, recall, and F1-score is less than $10^{-3}$.
    }
  \label{tab:classification}
\end{table*}

\begin{table*}[!h]
  \centering
    \begin{small}
        \input{tables/align_nile_results}
    \end{small}
  \caption{Alignment of automatic metrics with human preferences. Pairwise accuracy and $V$ are defined on [0:1] and $\kappa$ on [-1:1]. "w.g." indicates that the reference revision is provided. 
  Confidence interval of reported pairwise accuracy, $V$, and $\kappa$ score is less than $10^{-3}$.
  }
  \label{tab:pref_results}
\end{table*}

\begin{table*}[!ht]
  \centering
  \resizebox{\linewidth}{!}{
    \input{tables/runtimes/global_runtime}
    }
  \caption{
  GPU usage of all \ac{llmjudge} approaches on all three tasks using \phiquatre{} and \LlamaS.
  Confidence interval of reported times is less than $0.1$ minute.
  }
  \label{tab:consumption}
\end{table*}

\subsection{Baseline Metrics}

For all tasks, we include a set of similarity metrics that have been widely used for text generation: \Lev{} (normalised), \Bleu~\citep{papineni-etal-2002-bleu}, \Meteor~\citep{banerjee-lavie-2005-meteor} and \BertScore~\citep{zhang2020bertscore}.
We additionally use \textsc{Modern BERT}~\citep{warner-etal-2025-smarter}, a modernised encoder-only Transformer trained with updated data and techniques.
This yields an updated variant, \ModBScore{}, which we use as a \ac{SOTA} baseline.

We consider three \ac{llmjudge} variants: (a) LLM-Yes/No, which answers binary ``Yes/No'' questions, (b) \llmchoice{}, which conducts pairwise comparisons, and (c) \llmlikert{}, which assigns scores based on a five-point scale.
All prompts are provided in Appendix~\ref{apx:prompt_llmjudge}.

\subsection{*-PLUIE Metrics}

Prior work \citep{rios-kavuluru-2018-shot, nips_fslearners, JMLR:v25:23-0870, lemesle-etal-2025-paraphrase} shows that providing explicit examples of the task can improve the model’s ability to produce accurate and consistent judgement.
While \PPluie{} has demonstrated strong performance for evaluating text revision using its original paraphrase classification prompt~\citep{jourdan-etal-2025-identifying}, we investigate whether adapting the prompt to the task can further improve reliability and if it can be generalised to other tasks. 

To explore this hypothesis, we design personalised prompts for each studied task, considering \PPluie{} as a flexible plug-and-play metric that can easily be adapted through prompt and underlying perplexity model modification. 
We refer to the original paraphrase classification template as \tplPlui{Para} and construct the following variants:
\textbf{\tplPlui{Fr}}: a French adaptation of the original \tplPlui{Para} prompt, using translated few-shot examples to test French generalisation.
\textbf{\tplPlui{Net}}: a prompt for assessing whether two sentences express the same network policy.
Few-shot examples used are drawn and removed from the evaluation data to prevent bias.
\textbf{\tplPlui{Rev}}: a prompt designed to assess whether a generated revision follows its associated instruction. 
This variant uses the gold reference as a one-shot example to ground the model’s task understanding. 

All prompts are available in Appendix~\ref{apx:prompt_PLUIE}.
We compare \tplPlui{*} against the original \PPluie{} template to determine whether task-specific prompts improve alignment with human judgements. 
For all LLM-based methods, we use \phiquatreDef{}~\citep{abdinPhi4TechnicalReport2024}, and \LlamaSDef{}~\citep{urlllama} as the perplexity models.

\section{Results}

To study the natural interpretability of the proposed \tplPlui{*}, we compare approaches with standard similarity metrics and output-based \ac{llmjudge} methods.
As this metric grants a continuous score with an interpretable threshold, we use it for \textbf{classification} and for \textbf{preference evaluation}.
In classification, if the returned score is above $0$, the sample is considered as positive, and negative otherwise.
For preference, we rank the options according to their scores to identify the preferred one and align it to human preference.

\subsection{Classification}
\label{sec:res_classif}
We used \tplPlui{*} for classification of paraphrase/non-paraphrase pairs and good/bad Nile translations.
As baseline metric scores are not directly interpretable as categorical decisions, we calibrate each metric by determining an optimal decision threshold that maximises the F1 score.
Table~\ref{tab:classification} reports the performance of all evaluated metrics; we also report results for the natural threshold with \tplPlui{*}.

We find that \tplPlui{*/Para} achieves competitive or slightly better performance than \ac{llmjudge} metrics.
The difference between the calibrated and default thresholds is minimal, highlighting the interpretability and robustness of \tplPlui{*/Para}’s scoring scale.
Traditional metrics achieve an accuracy of approximately 33\% on French paraphrase classification, implying that all pairs are classified as paraphrases.
This finding suggests that the dataset is particularly challenging and that these surface-based metrics struggle to capture large semantic differences when lexical overlap remains high.

We only consider the Nile translations generated with Llama from the \ac{NEAT} methodology; other results can be found in Appendix~\ref{apx:Mistral-Ref}.
We observe that the \tplPlui{*/Para} consistently outperforms similarity metrics, and is comparable to output-based \ac{llmjudge}.
For this task, higher precision is preferable to avoid incorrect network configuration.
The best performance is observed with \tplPlui{Para} \LlamaS{} and the uncalibrated threshold ($0$).
Further analyses on the impact of the threshold choice are in Appendix~\ref{apx:fig_thresh_para}.

\subsection{Preference}

To assess the alignment between automatic metrics and human judgement, we use pairwise accuracy with tie calibration~\citep{deutsch2023ties}, Cramér’s $V$~\citep{cramer1946mathematical}, and Cohen’s $\kappa$~\citep{cohenkappa}; results are reported in Table~\ref{tab:pref_results}.

For Nile translation, all metrics correlate positively with human judgement, according to $\kappa$.
\tplPlui{Para} achieves the highest correlation, followed by \tplPlui{Net}. 
\ac{llmjudge} approaches are more dependent on the choice of the model, with \LlamaS\ emerging as the best option for this task.
\BertScore\ shows the best alignment among traditional similarity metrics, with scores comparable to the \llmlikert{} approach. 

For revision, across all three measures, \tplPlui{Rev} shows the highest alignment with human evaluations, achieving the best or second-best scores.
This surpasses other \ac{llmjudge} methods and traditional similarity metrics.
\llmchoice{} also performs competitively, while \llmlikert{} and \tplPlui{Para} achieve moderate yet consistent results.
In contrast, n-gram and embedding-based metrics display considerably weaker correlations with human judgement.
Lastly, for \tplPlui{Rev}, adding the reference revision in the prompt as a one-shot example does slightly improve the alignment judgement, making it the most reliable option to evaluate the task with or without a reference.
Results with different LLMs are available in Appendix~\ref{ap:LLMbased_metrics}, and show that \tplPlui{Rev} surpasses \llmchoice{} in all experimental configurations.




\subsection{Computational Efficiency}

We compare the computational cost of all LLM-based metrics in Table~\ref{tab:consumption}.
\tplPlui{*} approaches are consistently faster than output-based alternatives when using the same model, since they compute probabilities over limited answer tokens rather than generating long textual outputs.
Overall, \textsc{PLUIE} methods offer a favourable trade-off between efficiency and alignment, making them an attractive option for scalable \ac{llmjudge} evaluation.

As highlighted by \citet{nayab2025concisethoughtsimpactoutput}, the inference time of an LLM depends on the number of output tokens generated.
Due to the autoregressive nature of transformer decoders \citep{vaswaniAttentionAllYou}, each output token requires a dedicated forward pass, meaning that generation time scales linearly with the length of the response and is directly linked to the length of the input prompt.

Output-based LLM-judge methods are directly subject to this constraint: generating a free-form judgement of $\mathcal{N}(\hat{y})$ tokens requires $\mathcal{N}(\hat{y})$ successive decoder calls.
Furthermore, the prompt length $\mathcal{N}(x)$ itself is a source of overhead that is often overlooked.
Standard LLM-judge prompts tend to grow longer in practice~\cite{wang-etal-2025-wait}, as they must include explicit instructions describing the expected output format, and few-shot examples that condition the model to produce a structured response (\textit{e.g.}, a \textit{Yes}/\textit{No} answer or a Likert score) rather than free-form text.
Moreover, reasoning strategies such as Chains-of-Thought \cite{CoT-wei2022chain} further increase the length of generations.
Without such conditioning, output-based methods are prone to producing responses that are difficult to parse or do not conform to the expected schema.

\textsc{PLUIE} methods are free from this constraint: since the score is derived directly from the model's logits (as highlighted by Appendix~\ref{apx:pplui_def}), a well-formed confidence score is obtained regardless of how the model would have phrased its answer.
Explaining the response format or enforcing output structure through lengthy instructions is therefore not strictly necessary.
That said, providing few-shot examples remains beneficial: even though format enforcement is unnecessary, examples help realign the model's next-token distribution toward the tokens of interest (\textit{Yes} and \textit{No}), reducing the probability mass assigned to unrelated vocabulary entries and improving the discriminability of the resulting score.

\textsc{PLUIE} methods require \textbf{exactly one decoder pass}, regardless of the number of candidate tokens evaluated.
This follows from the fact that a single forward pass through the decoder produces a probability distribution over the entire vocabulary; the probabilities of \textit{Yes} and \textit{No} are therefore both available as entries of the same output logit vector, at no additional cost.
On the other hand, in the best-case scenario, by using an output-based LLM-judge method with the same prompt, the computational cost will be at least twice as expensive as a \textsc{PLUIE} method.
Indeed, even in the most favourable case, an output-based LLM-judge must produce at least two tokens, one for the answer and one end-of-sequence token, thereby requiring \textbf{at least two decoder passes}.
This explains the up to $7.9\times$ speedup reported in Table~\ref{tab:consumption}.

\input{reproducibility}

\section{Conclusion}

We introduced a generalised version of \PPluie{}, extending the original approach to a broader range of semantic evaluation tasks.
By adapting task-specific prompts, we showed that \tplPlui{*} consistently achieves stronger correlations with human judgement. 
Across all experiments, personalised \tplPlui{*} prompts are up to $7.9$ times faster than output-based approaches.
Its interpretable score and stable decision thresholds make it practical, avoid post-processing of LLM outputs and enable simple, scalable, and transparent model substitution for automatic LLM evaluation.

In addition, the proposed \tplPlui{Net} template can be used as a \ac{SOTA} alignment function between natural language and Nile intent expressions, as described in Appendix~\ref{apx:net_align_func}.
Overall, these results position \tplPlui{*} as an efficient and adaptable foundation for automatic evaluation in the era of LLMs.

\newpage
\section*{Limitations}

The perplexity models used with \tplPlui{*} were not fine-tuned for these experiments. 
Fine-tuning could potentially enhance both accuracy and task-specific sensitivity. 
The larger LLM used in this paper has 70B parameters; employing even larger models could further improve the results.

Most experiments were carried out in English, with additional tests in French showing consistent results.
The behaviour of the method in languages with richer morphology or markedly different syntactic structures remains to be investigated.

In this paper, we only considered prompts that can be formulated as ``Yes/No'' questions. 
It would be interesting to extend this work to several tasks that require different outputs. 
Such tasks could include sentiment analysis, topic classification or question answering.
Furthermore, the \tplPlui{*} formula could be changed to remove the one token limitation, which is discussed in Appendix \ref{apx:pplui_def}.

\section*{Ethical Considerations}

\textbf{Data Availability: }
For all considered tasks, we use datasets that are openly available. 
The datasets from~\citet{tytgat-etal-2024-evaluation} and~\citet{MUNSON2025} are available on demand to the original authors.
The paragraphs in ParaReval are extracted from scientific articles collected on OpenReview where they fall under different ``non-exclusive, perpetual, and royalty-free licence''.

\textbf{Computational Resources: }
Using LLMs remains resource-intensive; however, using an LLM of medium size like \phiquatreDef{} seems to be competitive with larger LLMs like \LlamaSDef{} (Tables \ref{tab:classification} and \ref{tab:pref_results}), and it's obviously less computationally intensive.
Lastly, \tplPlui{*}'s computational cost is much lower than other \ac{llmjudge} methods, as highlighted by Table \ref{tab:consumption}.

\section*{Acknowledgements}
\label{sec:thanks}

This work benefited from access to the MI300A resources of CINES under allocation 2025–AD011015262R1 granted by GENCI, and is supported by the Ministère des Armées - Agence de l'Innovation de la Défense.

This work was granted access to the HPC resources of IDRIS under the allocations 2023-AD011013901R1, 2024-AD011013901R2 and 2024-AD011014882R1 made by GENCI.

This study is partially funded by the ANR within the framework of the PIA EUR CyberSchool project (ANR-18-EURE-0004).


\bibliography{custom}

\appendix

\onecolumn

\section{\textrm{\PPluie} definition}
\label{apx:pplui_def}

\input{appendix/pluie_def}

\newpage

\section{French Paraphrase Dataset}
\label{apx:fr_para_dataset}

\input{appendix/tytgat_exemple}

\section{Example of data from ParaReval dataset}
~\label{apx:exp_revision}

\input{appendix/revision_example}

\section{\ac{llmjudge} prompts}
\label{apx:prompt_llmjudge}

\input{prompts/llmjudge}
\newpage

\section{*-PLUIE new task-specific prompts}
\label{apx:prompt_PLUIE}

In this section, we provide the prompts used for French paraphrase classification and Nile translation. For text revision, for \llmchoice{} and \llmlikert{}, we reused the prompts and generated evaluations from~\citet{jourdan-etal-2025-identifying}.

\input{prompts/instructpluie}

\newpage

\section{Mistral and Reference test results}
\label{apx:Mistral-Ref}

\input{tables/Nile_results_Mis-Ref}

\section{Score distribution according to the decision threshold}
\label{apx:fig_thresh_para}

\input{appendix/distrib_threshold}

\section{Additional results for alignment of LLM-based metrics on text revision}
\label{ap:LLMbased_metrics}

\input{tables/appendix_align_rev}

\section{\tplPlui{Net} as an alignment function}
\label{apx:net_align_func}

\input{appendix/netpluie}

\section{Additional results for French paraphrase classification}
\label{apx:add_french_llmjudge}

\input{tables/add_parap}
\input{appendix/additionnal_paraphrase}

\end{document}

%% file: tables/tytgat_distrib_table.tex
\begin{tabular}{l|ccccc|ccccc}
    \toprule
    \textbf{\textit{Task}} & \multicolumn{5}{c|}{\textbf{\textit{Paraphrase Classification}}} & \multicolumn{5}{c}{\textbf{\textit{Nile Translation}}}\\
    \textbf{Metric}&\textbf{Thr.}&\textbf{Acc.}&\textbf{Rec.}&\textbf{Prec.}&\textbf{F1} &\textbf{Thr.}&\textbf{Acc.}&\textbf{Rec.}&\textbf{Prec.}&\textbf{F1} \\ 
    \midrule
     \tplPlui{*} \phiquatre    & -7.63 & 0.71 & \underline{0.77} & 0.54 & \underline{0.63} &-3.14&\underline{0.81} &0.95 & 0.78 &0.85 \\
    \tplPlui{*} \phiquatre    & 0 & \underline{0.74} & 0.54 & 0.61 & 0.58 &0 &\underline{0.81} & 0.85 & 0.84 & 0.85 \\
    \tplPlui{*} \LlamaS       & -4.07 & 0.67 & 0.71 & 0.50 & 0.59 & 3.97 & 0.80 & 0.92 & 0.79 & 0.85 \\
    \tplPlui{*} \LlamaS       & 0 & 0.70 & 0.64 & 0.53 & 0.58 & 0 & 0.79 & 0.94 & 0.77 & 0.84 \\
    \midrule
    \tplPlui{Para} \phiquatre & -7.47 & 0.73 & 0.73 & 0.57 &\textbf{ 0.64} & -3.43&\underline{0.81} & 0.87 &0.83  &0.85\\
    \tplPlui{Para} \phiquatre & 0 & \textbf{0.75} & 0.49 & \textbf{0.66} & 0.56 &0&0.73 &0.63&\underline{0.88} &0.73 \\
    \tplPlui{Para} \LlamaS    & -11.91 & 0.70 & 0.66 & 0.54 & 0.59 & -13.73 & \underline{0.81} & 0.92 & 0.80  & 0.85\\
    \tplPlui{Para} \LlamaS    & 0 & 0.72 & 0.48 & 0.58 & 0.53 & 0 & \textbf{0.82} & 0.79 & \textbf{0.89}  & 0.84  \\
    \midrule

    LLM-Yes/No \phiquatre            &   &\underline{0.74} & 0.50 & \underline{0.64} & 0.56 &   & \underline{0.81} & 0.88 & 0.86 & \underline{0.87}\\

    LLM-Yes/No \LlamaS               &   & 0.71 & 0.48 & 0.56 & 0.52 &   & \underline{0.81} & 0.90 & 0.86 & \textbf{0.88}  \\

    \midrule
    \ModBScore{}            & 0.84 & 0.33 & \textbf{1.00} & 0.33 & 0.49 &0.68&0.63 &\underline{0.96} & 0.63 & 0.76 \\
    \BertScore{}            & 0.80 & 0.33 & \textbf{1.00} & 0.33 & 0.49 &0.39&0.67 &\textbf{1.0} &0.64&0.78  \\
    \midrule
    \Meteor{}               & 0.43 & 0.33 & \textbf{1.00} & 0.33 & 0.49 &0.0&0.6 &\textbf{1.0} & 0.6 & 0.75 \\
    \Bleu{}                 & 0.00 & 0.33 & \textbf{1.00} & 0.33 & 0.49 &0.0&0.6 &\textbf{1.0} &0.6 &0.75 \\
    \Lev{}                  & 0.71 & 0.33 & \textbf{1.00} & 0.33 & 0.49 &0.17&0.65 & 0.95 &0.64&0.77  \\
    \midrule
    \textit{Random weighted} &   & \textit{0.56} & \textit{0.32} & \textit{0.32} & \textit{0.32} &   & 0.52 & 0.60 & 0.61 & 0.61\\
    \textit{Random uniform} &   & \textit{0.50} & \textit{0.50} & \textit{0.32} & \textit{0.40}&   & 0.49 & 0.48 & 0.60 & 0.53\\
    \bottomrule

\end{tabular}

%% file: tables/align_nile_results.tex
\begin{tabular}{ l| ccc | cccccc}
    \toprule
   
    \textbf{\textit{Task}}  & \multicolumn{3}{c|}{\textit{\textbf{Nile Translation}}} & \multicolumn{6}{c}{\textit{\textbf{Scientific Text Revision}}}\\
    \textbf{Metric} & \textbf{Pair acc.} &\textbf{$V$} & \textbf{$\kappa$} & \multicolumn{2}{c}{\textbf{Pair acc.}} & \multicolumn{2}{c}{\textbf{$V$}} & \multicolumn{2}{c}{\textbf{$\kappa$}}  \\
    
    \midrule
    
    \tplPlui{*} \phiquatre  & 0.69 & 0.42 & 0.40  & \textbf{0.61} & \underline{0.61} {\scriptsize w.g.} & \textbf{0.31} & \textbf{0.32} {\scriptsize w.g.} & \textbf{0.32} & \underline{0.33} {\scriptsize w.g.}\\

    \tplPlui{*} \LlamaS     & \underline{0.70} & 0.43 & 0.42  & \textbf{0.61} & \textbf{0.62} {\scriptsize w.g.} & \textbf{0.31} & \textbf{0.32} {\scriptsize w.g.} & \textbf{0.32}& \textbf{0.34} {\scriptsize w.g.}\\
    
    \tplPlui{Para} \phiquatre & \underline{0.70} & \underline{0.44} & 0.42  &  \multicolumn{2}{c}{0.52} & \multicolumn{2}{c}{0.21} & \multicolumn{2}{c}{0.15}\\
    
    \tplPlui{Para} \LlamaS    & \textbf{0.72} & \textbf{0.46} & \underline{0.43}   & \multicolumn{2}{c}{0.52} & \multicolumn{2}{c}{0.20} & \multicolumn{2}{c}{0.17} \\ 
    \midrule
    
    \llmchoice{}  \phiquatre & 0.47 & 0.42 & 0.20 & 0.53 & 0.55 {\scriptsize w.g.} & 0.25 & 0.27 {\scriptsize w.g.} & 0.24 & 0.27 {\scriptsize w.g.}\\
    
    \llmchoice{} \LlamaS             & 0.47 & 0.43 & 0.20 & \underline{0.59} & 0.60 {\scriptsize w.g.} & 0.28 & \underline{0.30} {\scriptsize w.g.} & \underline{0.30} & 0.31 {\scriptsize w.g.}\\
    
    \llmlikert{} \phiquatre          & 0.65 & 0.37 & 0.41 & 0.45 & 0.52 {\scriptsize w.g.} & 0.30 & 0.29 {\scriptsize w.g.} & 0.21 & 0.26 {\scriptsize w.g.}\\
    \llmlikert{} \LlamaS             & 0.69 & 0.40 & \textbf{0.44} &0.44 & 0.50 {\scriptsize w.g.} & \underline{0.29} & 0.27 {\scriptsize w.g.} & 0.19 & 0.23 {\scriptsize w.g.}  \\
    \midrule
    
    \ModBScore                     & 0.63 & 0.41 & 0.36 & \multicolumn{2}{c}{0.36} & \multicolumn{2}{c}{0.13} & \multicolumn{2}{c}{-0.07} \\
    \BertScore                     & 0.68 & 0.43 & 0.40 & \multicolumn{2}{c}{0.45} & \multicolumn{2}{c}{0.16} & \multicolumn{2}{c}{0.03}  \\ 
    \midrule
    \Meteor                        & 0.68 & 0.40 & 0.37 &  \multicolumn{2}{c}{0.42} & \multicolumn{2}{c}{0.19} & \multicolumn{2}{c}{0.00}\\ 
    \Bleu                          & 0.43 & N/A & 0.00 & \multicolumn{2}{c}{0.41} & \multicolumn{2}{c}{0.17} & \multicolumn{2}{c}{-0.03}\\
    \Lev                           & 0.64 & 0.34 & 0.29 &  \multicolumn{2}{c}{0.44} & \multicolumn{2}{c}{0.13} & \multicolumn{2}{c}{0.01}  \\
    \midrule
    \textit{Random weighted} & \textit{0.43} & \textit{0.06}& \textit{0.00} & \multicolumn{2}{c}{\textit{0.36}}   & \multicolumn{2}{c}{\textit{0.03}}    & \multicolumn{2}{c}{\textit{0.00}}\\
    \textit{Random uniform} & \textit{0.33} & \textit{0.06} & \textit{0.00} & \multicolumn{2}{c}{\textit{0.33} }  & \multicolumn{2}{c}{\textit{0.03}}    & \multicolumn{2}{c}{\textit{0.00} } \\
    \bottomrule
\end{tabular}  

%% file: tables/runtimes/global_runtime.tex
 \begin{tabular}{l | l|c c| cc | ccc}
    \toprule
    
    & \textbf{\textit{Task}} &\multicolumn{2}{c|}{\textbf{\textit{Paraphrase Classification}}} & \multicolumn{2}{c|}{\textbf{\textit{Nile Translation}}} & \multicolumn{3}{c}{\textbf{\textit{Scientific Text Revision}}} \\
    
    \textbf{model} & \textbf{Approach} & \textbf{GPUs} & \textbf{Runtime} & \textbf{GPUs} & \textbf{Runtime} & \textbf{GPUs} & \multicolumn{2}{c}{\textbf{Runtime}} \\
    \midrule
    \parbox[t]{0mm}{\multirow{4}{*}{\rotatebox[origin=c]{0}{\phiquatre} }} 
    & 
      LLM-Yes/No / \llmchoice{}          & MI300 x1 & 23 min & MI300 x1 & 11 min & A100 x1 & 40 min & 41 min {\scriptsize w.g.}\\
      
    & \llmlikert{}  & &  & MI300 x1 & 10 min & A100 x1 & 51 min & 65 min {\scriptsize w.g.}\\

    & \tplPlui{Para}   & MI300 x1 & \textbf{3.5 min} & MI300 x1 & \underline{1.8 min} & A100 x1 & \textbf{08 min} & \\
    
    & \tplPlui{*} & MI300 x1 & \underline{3.6 min} & MI300 x1 & \textbf{1.4 min} & A100 x1 & \underline{09 min} & 13 min {\scriptsize w.g.}\\
    \midrule

    \parbox[t]{0mm}{\multirow{4}{*}{\rotatebox[origin=c]{0}{{\LlamaS}} }} 
    & 
    LLM-Yes/No / \llmchoice{}   & MI300 x2 & 48 min  & MI300 x2 & 22 min & A100 x2 & 124 min & 127 min {\scriptsize w.g.} \\
    
    & \llmlikert{} &   &  & MI300 x2 & 21 min & A100 x2 & 141 min & 187 min {\scriptsize w.g.} \\

    & \tplPlui{Para}  & MI300 x2 & \textbf{14 min} & MI300 x2 & \underline{6.5 min} & A100 x3 & \textbf{29 min} &\\
    
    & \tplPlui{*}     & MI300 x2 & \underline{17 min} & MI300 x2 & \textbf{3.4 min} & A100 x3 & \underline{33 min} & 55 min {\scriptsize w.g.}\\
    
    \bottomrule
  \end{tabular} 

%% file: reproducibility.tex










\section{Reproducibility}

To use \tplPlui{*} you can follow the documentation available on a \href{https://huggingface.co/spaces/qlemesle/parapluie}{HuggingFace space (https://huggingface.co/spaces/qlemesle/parapluie)} and use the publicly available source code on \href{https://gitlab.inria.fr/expression/paraphrase-generation-evaluation-powered-by-an-llm-a-semantic-metric-not-a-lexical-one-coling-2025}{GitLab (https://gitlab.inria.fr/expression/paraphrase-generation-evaluation-powered-by-an-llm-a-semantic-metric-not-a-lexical-one-coling-2025)}.

Both the underlying LLM and the prompting template can be swapped independently without modifying the scoring logic.
The following snippet illustrates how to instantiate and run a \tplPlui{*} variant:

\begin{pycode}
from PPLUIE.wrapper import ParaPLUIE

scorer = ParaPLUIE()
scorer.init("microsoft/phi-4", device="auto")
# Use Fr-PLUIE by simply switching the template
scorer.setTemplate("FS-DIRECT_FR")

S = ["Les enfants ont boulonn(*@\'e@*) tous les g(*@\^a@*)teaux."]
H = ["Les enfants ont mang(*@\'e@*) tous les g(*@\^a@*)teaux."]

score = scorer.compute(S, H)
print("Result score : ", score) # -4.85
# score > 0 : paraphrase, score < 0 : non-paraphrase
\end{pycode}

The full list of supported models and templates is available via \texttt{scorer.show\_available\_models()} and \texttt{scorer.show\_templates()}.
New task-specific templates can be easily added by extending the \texttt{template.py} file.

%% file: appendix/pluie_def.tex
\begin{figure}[H]
    \centering
    \includegraphics[width=.8\linewidth]{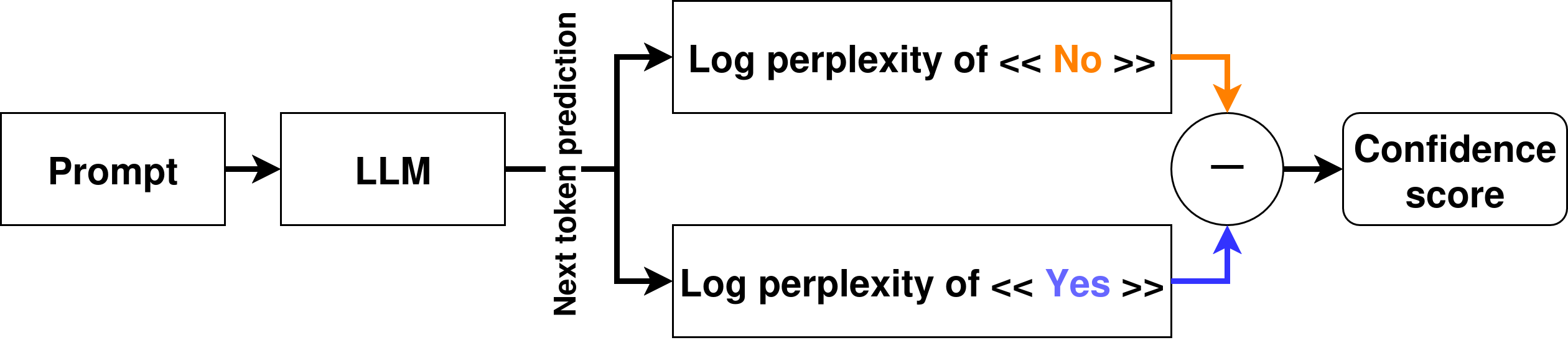}
\caption{\tplPlui{*} workflow.}
\label{fig:plui-workflow}
\end{figure}

\textrm{\PPluie} is originally defined as the log-likelihood ratio to compare the predominance of an answer “Yes” against “No” to the question raised by the prompt~\citep{lemesle-etal-2025-paraphrase}, as illustrated in Figure~\ref{fig:plui-workflow}.

More formally, given:
\begin{itemize}\setlength\itemsep{-1pt}
    \item $S$ the source sentence,
    \item $H$ the hypothetical sentence,
    \item $\Prompt{}(S,H)$ the prompt filled with the sentences $S$ and $H$,
    \item $T$: length in tokens of the $\Prompt{}(S,H)$,
    \item $\circ$: the concatenation operator of 2 sequences of tokens.
\end{itemize}

\begin{eqnarray}
      \texttt{\PPluie{}}(S,H) &=& \log\left(\frac{p\left(\textrm{Yes}|\Prompt{}(S,H)\right)}{p\left(\textrm{No}|\Prompt{}(S,H)\right)}\right) \label{eq:para-1} 
\label{equ:pplui-proof}
\end{eqnarray}

It can be expanded, using the Bayes formula, to bring out the perplexity of the template:

\begin{eqnarray}
    & = & \log \left(\frac{p\left(\Prompt{}(S,H)\circ\textrm{Yes}\right)}{p\left(\Prompt{}(S,H)\right)} \times \frac{p\left(\Prompt{}(S,H)\right)}{p\left(\Prompt{}(S,H)\circ\textrm{No}\right)}\right) \nonumber\\
    & = & - \frac{-1}{T+1}\log \left(p\left(\Prompt{}(S,H)\circ\textrm{Yes}\right)\right)\times(T+1) \nonumber \\ 
    & & + \frac{-1}{T+1} \log \left(p\left(\Prompt{}(S,H)\circ\textrm{No}\right)\right)\times(T+1) \nonumber\\ 
    & = & \log\left(\exp\left(\frac{-1}{T+1}\log \left(p\left(\Prompt{}(S,H)\circ\textrm{No}\right)\right)\right)^{T+1}\right) \nonumber\\
    & & - \log\left(\exp\left(\frac{-1}{T+1} \log \left(p\left(\Prompt{}(S,H)\circ\textrm{Yes}\right)\right)\right)^{T+1}\right) \nonumber\\
    & = & \log\left(ppl\left(\Prompt{}(S,H)\circ\textrm{No}\right)^{T+1}\right) \nonumber \\
    & & - \log\left(ppl\left(\Prompt{}(S,H)\circ\textrm{Yes}\right)^{T+1}\right) \label{eq:tplusone} \\
    & = & (T+1) \times \log\left(ppl\left(\Prompt{}(S,H)\circ\textrm{No}\right)\right) \nonumber\\
    & & - (T+1) \times \log\left(ppl\left(\Prompt{}(S,H)\circ\textrm{Yes}\right)\right) \nonumber
\end{eqnarray}

Finally, assuming that the \llm{} employed uses log perplexity as a \loss{}, we have:

\begin{align}
    & \texttt{\PPluie{}}(S,H) =  (T+1) \times \left[\textrm{$loss$}_{\llm{}}(\Prompt{}(S,H) \circ \textrm{No})  -  \textrm{$loss$}_{\llm{}}(\Prompt{}(S,H) \circ \textrm{Yes})\right]
\end{align}

Note that \PPluie{} is not necessarily symmetrical.
\PPluie{} presents plug-and-play capabilities: both the underlying perplexity model and the prompt template can be modified without altering the fundamental principle of the method. 
However, a few aspects must be taken into consideration.
First, the perplexity model must be able to perform question-answering tasks, as the metric relies on evaluating the model’s confidence between two mutually exclusive answers.
Note that a requirement of Equation~\ref{eq:tplusone} is that the two answers being opposed each correspond to exactly one token according to the model’s tokenizer.
Beyond that, the choice of tokens is not restricted (it could be “Dog” vs. “Cat”), even though we only consider “Yes” vs. “No” in this study.
Second, to ensure correct perplexity computation, the tokens being compared must appear as the final tokens of the prompt.
Tokenizers of most modern LLM-based chatbots employ a user–assistant dialogue format, where special tokens are inserted to mark the end of each role.
These role-ending tokens must be removed prior to perplexity computation, as they would otherwise alter the results.
For example, when computing the perplexity of “Yes” in the following dialogue:

\texttt{<user>People like “Cats” more than “Dogs”.</user> <assistant> Yes </assistant>} the end-of-turn token \texttt{</assistant>} must be removed before calculating perplexity.
We rely on the publicly available implementation provided by the original authors\footnote{\footnotesize{\href{https://gitlab.inria.fr/expression/paraphrase-generation-evaluation-powered-by-an-llm-a-semantic-metric-not-a-lexical-one-coling-2025}{\texttt{https://gitlab.inria.fr/expression/paraphrase\allowbreak{}-generation-\allowbreak{}evaluation-\allowbreak{}powered-\allowbreak{}by-\allowbreak{}an-llm-\allowbreak{}a-\allowbreak{}semantic\allowbreak{}-metric\allowbreak{}-not-a-lexical-one-coli</assistant>ng-2025}}}}.

This formula can be generalised to answers of different length. 
So, if one answer is ($Y_1$\dots\ $Y_i$) and the second is ($N_1$\dots\ $N_j$), we can write:

{\small
\begin{align}
    \texttt{\PPluie{}}(S,H) =  (T+j) \times \left(\textrm{$loss$}_{\llm{}}(\Prompt{}(S,H) \circ N_1\dots{}N_j)  - (T+i) \times \textrm{$loss$}_{\llm{}}(\Prompt{}(S,H) \circ Y_1\dots{}Y_i)\right)
\end{align}}%

A drawback here is that the shorter answer would be favoured: indeed, it is often more probable to generate a short sequence rather than a long one.
Considering the average perplexity of the answer could be an alternative, but we would likely observe a trend of centering every answer around a similar average, which would be regrettable.

%% file: appendix/tytgat_exemple.tex






\begin{table}[!h]
\centering
\begin{tabular}{l|c}
\toprule
\textbf{Transformation}&\textbf{Sentence}\\
\midrule

None               & Les enfants ont boulotté tous les gâteaux. \\
                   & \textit{The children gobbled up all the cakes.} \\
                   \midrule
Paronym            & Les enfants ont boulonné tous les gâteaux \\
                   & \textit{The children screwed all the cakes.} \\
                   \midrule
Synonym            & Les enfants ont mangé tous les gâteaux. \\
                   & \textit{The children ate all the cakes.} \\
                   \midrule
Synonym of Paronym & Les enfants ont fixé tous les gâteaux. \\
                   & \textit{The children fixed all the cakes.} \\
\bottomrule

\end{tabular}

\caption{
    Characteristic sentence of \citet{tytgat-etal-2024-evaluation} dataset and its different transformations.
    English translations are provided in \textit{italics}; however, as paronym substitution relies on visual and phonetic similarity, it cannot be easily translated. 
    A simple example of an English paronym pair:
    ``The children decided to eat the \textbf{desert}.'' vs.\
    ``The children decided to eat the \textbf{dessert}.''
}

\label{tab:example_tytgat}
\end{table}

This dataset is built on the work of~\citet{tytgat-etal-2024-evaluation} who showed that conventionally used semantic similarity metrics are often more sensitive to surface-level differences than to semantic variations.
Their study introduced an expert-annotated French dataset of 355 source sentences, each constructed independently of any specific domain.
For each original sentence,~\citet{tytgat-etal-2024-evaluation} identified the central semantic word, the one that most conveys the meaning of the sentence, and systematically produced three types of modified versions:
\begin{itemize}
    \item \textbf{Paronym substitution}: the word is replaced by a paronym, a word that looks or sounds similar but has a completely different meaning.
The resulting sentence becomes semantically different from the original while remaining lexically close.
\item \textbf{Synonym substitution}: the word is replaced by a synonym, producing a sentence that preserves the original meaning while being lexically close.
\item \textbf{Synonym of a paronym}: the word is replaced by a synonym of the paronym, leading to an additional nuanced variant.
This process results in three alternative versions for each source sentence, an example of these transformations is provided in Table~\ref{tab:example_tytgat}.
\end{itemize}

Let us denote the transformation of a sentence \textbf{s} by synonym \textbf{S} and by paronym \textbf{P}.
We create a new set composed of pairs of sentences formed by the source sentence $s$ and its transformation.
Depending on the combination of transformations, we label them as paraphrase or non-paraphrase:

$ \textbf{( s, S(s) )} \implies \text{paraphrase} $
    
$ \textbf{( P(s), S(P(s)) )} \implies \text{paraphrase} $
    
$ \textbf{( s, P(s) )} \implies \overline{\text{paraphrase}} $   
    
$ \textbf{( s, S(P(s)) )} \implies \overline{\text{paraphrase}} $
    
$ \textbf{( P(s), S(s) )} \implies \overline{\text{paraphrase}} $

$ \textbf{( S(s), S(P(s)) )} \implies \overline{\text{paraphrase}} $

~

The resulting dataset is composed of $1914$ pairs with $33.60$\% positive pairs.

%% file: appendix/revision_example.tex
This dataset focuses on the paragraph-level scientific text revision task, where a paragraph and an accompanying instruction specifying the intended modification are provided as input, and the model is expected to produce a revised paragraph that aligns with the given instruction~\citep{jourdan-etal-2025-pararev}.

We use the test split of the ParaReval  dataset\footnote{\url{https://github.com/JourdanL/parareval}}~\citep{jourdan-etal-2025-identifying}, a collection of human pairwise evaluations of automatically generated revisions. 
Each instance in the dataset consists of two versions of the same paragraph, extracted from computer science papers from OpenReview and authored by the original writers. 
Pairs of paragraphs are annotated with an instruction describing the underlying revision intention; an example is given in Appendix~\ref{apx:exp_revision}. 

Based on these annotations, each paragraph was automatically revised by six different models following the corresponding instruction. 
The resulting generated revisions were evaluated in pairs by human annotators to assess the reliability of automatic evaluation metrics for this task. 
Annotators answered three questions: (1) Did the model follow the instruction? (2) Is the revision correct, \textit{i.e.}, better than the original? and (3) Which revision do they prefer between options A and B? The annotator's responses enable ranking the two options and identifying the preferred one. 

The data comprises 258 pairs of revised paragraphs, each annotated with two distinct revision instructions, resulting in a total of 516 evaluation instances.
Human pairwise preferences are balanced, with annotators favouring option A in 44\% of cases, option B in 41\%, and reporting ties in 15\% of cases.

\begin{figure}[H]
    \centering
    \includegraphics[width=0.99\textwidth]{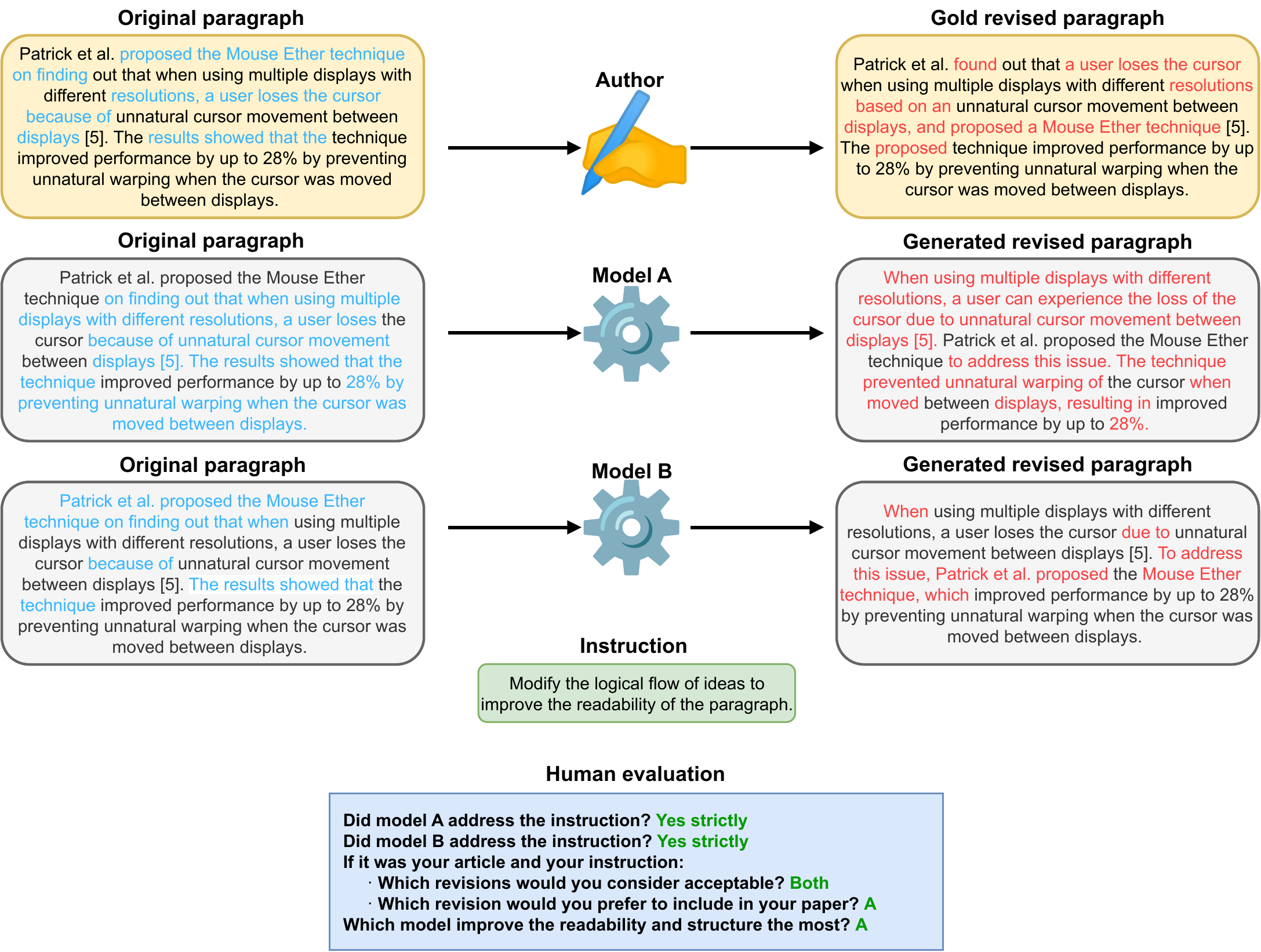}
\caption{Example of data in the ParaReval dataset}
\label{fig:example_pararev}
\end{figure}

%% file: prompts/llmjudge.tex
\begin{codelisting} {Prompt for LLM-Yes/No for paraphrase detection}
system="""(*@You will receive two sentences A and B, you will have to identify if they mean the same thing.
 In your answer please only provide the answers to the question.@*)"""

user="""[BEGIN EXAMPLES]
***
[Sentence A]: Amrozi accused his brother , whom he called  the witness  , of deliberately distorting his evidence .
[Sentence B]: Amrozi accused his brother , whom he disparagingly referred to as 'the liar witness', of intentionally twisting his testimony .
No
***
[Sentence A]: Pennmakkal is an Indian Malayalam film from 1966 , produced by J. Sasikumar and directed by KP Kottarakkara .
[Sentence B]: The Indian Malayalam film 'Pennmakkal', released in 1966, was produced by J. Sasikumar and directed by KP Kottarakkara .
Yes
***
[Sentence A]: Sorkin , who faces charges of conspiracy to obstruct justice and lying to a grand jury , was to have been tried separately .
[Sentence B]: Despite being accused of conspiring to obstruct justice and perjury , Sorkin was supposed to stand trial on his own .
No
***
[Sentence A]: Gilroy police and FBI agents described Gehring as cooperative , but said Saturday that he had revealed nothing about what had happened to the children .
[Sentence B]: Although Gilroy police and FBI agents reported that Gehring was cooperative , he hadn't disclosed any information about the children's whereabouts or what had happened to them as of Saturday .
No
***
[Sentence A]: Whereas '' e `` the electric charge of the particle and A is the magnetic vector potential of the electromagnetic field .
[Sentence B]: The electric charge of the particle is denoted by ''e'', and the magnetic vector potential of the electromagnetic field is denoted by 'A' .
Yes
***
[Sentence A]: The Jidanul River is a tributary of the Jiul de Vest River in Romania .
[Sentence B]: The Jidanul River is a mere insignificant stream that flows into the grand Jiul de Vest River in Romania .
No
***
[END EXAMPLES]

[BEGIN DATA]
[Sentence A]: "(*@\textcolor{blue}{\{source\}}@*)"
***
[Sentence B]: "(*@\textcolor{blue}{\{paraphrase\}}@*)"
***
[END DATA]

Do these two sentences express the same meaning? Answer "Yes" or "No". 

"""+"""
You do not need to explain the reason.

Your response must be RFC8259 compliant JSON following this schema: 
{{"answer": str }}"""
\end{codelisting}

To evaluate in similar settings the \tplPlui{Fr} and LLM-Yes/No methods, we design a French prompt for them. Results with this prompt are available in Appendix~\ref{apx:add_french_llmjudge}.

\begin{codelisting} {French variant of prompt for LLM-Yes/No for paraphrase detection}
system="""(*@Tu vas recevoir deux phrases, A et B, tu vas devoir identifier si elles signifient la m\^eme chose.
 Dans ta r\'eponse fournis uniquement la r\'eponse \`a la question.@*)"""

user="""[DEBUT EXEMPLES]
***
(*@[Phrase A]: Amrozi a accus\'e son fr\`ere, qu'il appelait le t\'emoin, d'avoir d\'elib\'er\'ement d\'eform\'e ses preuves .@*)
(*@[Phrase B]: Amrozi a accus\'e son fr\`ere, qu'il d\'esignait de mani\`ere p\'ejorative comme le t\'emoin menteur, d'avoir intentionnellement falsifi\'e son t\'emoignage.@*)
Non
***
(*@[Phrase A]: Pennmakkal est un film indien en malayalam de 1966, produit par J. Sasikumar et r\'ealis\'e par KP Kottarakkara .@*)
(*@[Phrase B]: Le film indien en malayalam 'Pennmakkal', sorti en 1966, a \'et\'e produit par J. Sasikumar et r\'ealis\'e par KP Kottarakkara .@*)
Oui
***
(*@[Phrase A]: Sorkin, qui fait face \`a des accusations de complot pour entraver la justice et de faux t\'emoignage devant un grand jury, devait \^etre jug\'e s\'epar\'ement .@*)
(*@[Phrase B]: Malgr\'e les accusations de complot pour entraver la justice et de parjure, Sorkin devait \^etre jug\'e seul .@*)
Non
***
(*@[Phrase A]: La police de Gilroy et les agents du FBI ont d\'ecrit Gehring comme coop\'eratif, mais ont d\'eclar\'e samedi qu'il n'avait r\'ev\'el\'e aucune information sur ce qui \'etait arriv\'e aux enfants.@*)
(*@[Phrase B]: Bien que la police de Gilroy et les agents du FBI aient rapport\'e que Gehring \'etait coop\'eratif, il n'avait pas divulgu\'e d'informations sur le lieu o\`u se trouvaient les enfants ou sur ce qui leur \'etait arriv\'e samedi .@*)
Non
***
(*@[Phrase A]: Dans lequel ''e'' repr\'esente la charge \'electrique de la particule et A est le vecteur du potentiel magn\'etique du champ \'electromagn\'etique .@*)
(*@[Phrase B]: La charge \'electrique de la particule est d\'esign\'ee par ''e'', et le vecteur du potentiel magn\'etique du champ \'electromagn\'etique est d\'esign\'e par 'A' .@*)
Oui
***
(*@[Phrase A]: La rivi\`ere Jidanul est un affluent de la rivi\`ere Jiul de Vest en Roumanie .@*)
(*@[Phrase B]: La rivi\`ere Jidanul est un simple ruisseau insignifiant qui se jette dans la grande rivi\`ere Jiul de Vest en Roumanie .@*)
Non
***
[FIN EXEMPLES]

[DEBUT DONNEES]
(*@[Phrase A]: "\textcolor{blue}{\{source\}}@*)"
***
(*@[Phrase B]: "\textcolor{blue}{\{paraphrase\}}@*)"
***
[FIN DONNEES]

(*@Est-ce que ces deux phrases veulent dire la m\^eme chose ? R\'eponds par "Oui" ou "Non".@*)

"""+"""
(*@Tu n'as pas besoin d'expliquer la raison.@*)

(*@Ta r\'eponse doit \^etre compatible RFC8259 JSON et suivre le sch\'ema suivant : @*)
{{"(*@r\'eponse@*)": str }}"""
\end{codelisting}

\begin{codelisting} {Prompt for LLM-Yes/No for Nile translation}

system="""(*@You are an evaluator of nile network policies to english translations. In this task, a translation model has been provided with the original nile sentence and translated it to english. You will be given the original nile sentence and the proposition from this model and will have to report if the two sentences express the same network policy. In your answer please only provide the answers to the questions.@*)"""

user="""[BEGIN EXAMPLES]
***
[Sentence A]: Everquest is blocked by the University firewall
[Sentence B]: Everquest is not allowed by the University firewall
Yes
***
[Sentence A]: Quotas for students are 5000 Megabyte per hr download and 2000 Megabyte per hour upload
[Sentence B]: Students have a download quota of 5000 MB per hour and an upload quota of 2000 MB per hour
Yes
***
[Sentence A]: from endpoint('guests') to endpoint('servers') for group('host') block traffic('any')
[Sentence B]: RHIT-OPEN will work in labs, classrooms and residence halls but does not allow the user to access the internal Rose-Hulman network
No
***
[Sentence A]: Housing does not normally limit the amount of bandwidth
[Sentence B]: There is no bandwidth limit for the dorms
No
***
[Sentence A]: for endpoint('university') add middlebox('firewall') allow traffic('H323 video conferencing')
[Sentence B]: H323 video conferencing is allowed by the University firewall
Yes
***
[Sentence A]: H323 video conferencing is allowed by the University firewall
[Sentence B]: The network firewall does not block H.323 video conferencing
No
***
[END EXAMPLES]

[BEGIN DATA]
[Sentence A]: "(*@\textcolor{blue}{\{nile\}}@*)"
***
[Sentence B]: "(*@\textcolor{blue}{\{translation\}}@*)"
***
[END DATA]

Do these two sentences express the same network policy? Answer "Yes" or "No". 

"""+"""
You do not need to explain the reason.

Your response must be RFC8259 compliant JSON following this schema: 
{{"answer": str }}"""

\end{codelisting}

\begin{codelisting} {Prompt for LLM-choice for Nile translation}
system="""You are an evaluator of nile network policies to english translations. In this task, two translation models have been provided with the original nile sentence and translated it to english. You will be given the proposition from the two different models and a question to identify the best one. In your answer please only provide the answers to the questions."""

user="""[BEGIN DATA]
***
[Nile network policy]: "(*@\textcolor{blue}{\{nile\}}@*)"
***
[Sentence A]: "(*@\textcolor{blue}{\{translation 1\}}@*)"
***
[Sentence B]: "(*@\textcolor{blue}{\{translation 2\}}@*)"
***
[END DATA]

Which sentence is the best translation of the nile network policy? Answer "A", "B" or "Tie". 

"""+"""
You do not need to explain the reason.

Your response must be RFC8259 compliant JSON following this schema: 
{{"best_translation": str }}"""
\end{codelisting}

\begin{codelisting} {Prompt for LLM-Likert for Nile translation}
system="""You are an evaluator of nile network policies to english translations. 
In this task, a translation model has been provided with the original nile sentence and translated it to english.
You will be given the proposition from the translation model and an affirmation to determine the quality of this proposition.
You will answer this affirmation with a grade (int) from 1 to 5 as follows: 1 = Strongly disagree , 2 = Disagree , 3 = Neutral , 4 = Agree , 5 = Strongly agree 
To help you in this task, you will also be given the gold reference which is a correct translation to english.
In your answer please only provide the answer to the affirmation."""

user="""[BEGIN DATA]
***
[Nile network policy]: "(*@\textcolor{blue}{\{nile\}}@*)"
***
[Model proposed translation]: "(*@\textcolor{blue}{\{translation\}}@*)"
***
[END DATA]

The model proposed a correct translation of the nile network policy?

"""+"""
You do not need to explain the reason.

Your response must be RFC8259 compliant JSON following this schema: 
{{"translation_quality": int }}"""
\end{codelisting}

%% file: prompts/instructpluie.tex
\begin{codelisting} {\tplPlui{French} new template for French paraphrase detection}     
user= """(*@Tu vas recevoir deux phrases, A et B. Est-ce que ces deux phrases veulent dire la même chose ? Réponds avec un seul mot : "oui" ou "non".@*)"""

assistant="""(*@S'il te plaît, fournis-moi les deux phrases que je dois évaluer.@*)"""

user="""A: (*@"Amrozi a accusé son frère, qu'il appelait "le témoin", d'avoir délibérément déformé ses preuves ."; B: "Amrozi a accusé son frère, qu'il désignait de manière péjorative comme "le témoin menteur", d'avoir intentionnellement falsifié son témoignage."@*)"""

assistant="""non"""

user="""A: (*@"Pennmakkal est un film indien en malayalam de 1966, produit par J. Sasikumar et réalisé par KP Kottarakkara ."; B: "Le film indien en malayalam 'Pennmakkal', sorti en 1966, a été produit par J. Sasikumar et réalisé par KP Kottarakkara ."@*)"""

assistant="""oui"""

user="""A: (*@"Sorkin, qui fait face à des accusations de complot pour entraver la justice et de faux témoignage devant un grand jury, devait être jugé séparément ."; B: "Malgré les accusations de complot pour entraver la justice et de parjure, Sorkin devait être jugé seul ."@*)"""

assistant="""non"""

user="""A: (*@"La police de Gilroy et les agents du FBI ont décrit Gehring comme coopératif, mais ont déclaré samedi qu'il n'avait révélé aucune information sur ce qui était arrivé aux enfants ."; B: "Bien que la police de Gilroy et les agents du FBI aient rapporté que Gehring était coopératif, il n'avait pas divulgué d'informations sur le lieu où se trouvaient les enfants ou sur ce qui leur était arrivé samedi ."@*)"""

assistant="""non"""

user="""A: (*@"Dans lequel ''e'' représente la charge électrique de la particule et A est le vecteur du potentiel magnétique du champ électromagnétique ."; B: "La charge électrique de la particule est désignée par ''e'', et le vecteur du potentiel magnétique du champ électromagnétique est désigné par 'A' ."@*)"""

assistant="""oui"""

user="""A: (*@"La rivière Jidanul est un affluent de la rivière Jiul de Vest en Roumanie ."; B: "La rivière Jidanul est un simple ruisseau insignifiant qui se jette dans la grande rivière Jiul de Vest en Roumanie ."@*)"""

assistant="""non"""

user="""A: "(*@\textcolor{blue}{\{source\}}@*)"; B: "(*@\textcolor{blue}{\{paraphrase\}}@*)""""

\end{codelisting}

\begin{codelisting} {\tplPlui{Net} new template for network policy}     
user= """You will receive two sentences A and B. Do these two sentences express the same network policy? Answer with only one word "Yes" or "No"."""

assistant= """Please provide the data for me to evaluate."""

user="""A: "Everquest is blocked by the University firewall"; B: "Everquest is not allowed by the University firewall""""

assistant="""Yes"""

user="""A: "Quotas for students are 5000 Megabyte per hr download and 2000 Megabyte per hour upload"; B: "Students have a download quota of 5000 MB per hour and an upload quota of 2000 MB per hour""""

assistant="""Yes"""

user="""A: "from endpoint('guests') to endpoint('servers') for group('host') block traffic('any')"; B: "RHIT-OPEN will work in labs, classrooms and residence halls but does not allow the user to access the internal Rose-Hulman network""""

assistant="""No"""

user="""A: "Housing does not normally limit the amount of bandwidth"; B: "There is no bandwidth limit for the dorms""""

assistant="""No"""

user="""A: "for endpoint('university') add middlebox('firewall') allow traffic('H323 video conferencing')"; B: "H323 video conferencing is allowed by the University firewall""""

assistant="""Yes"""

user="""A: "H323 video conferencing is allowed by the University firewall"; B: "The network firewall does not block H.323 video conferencing""""

assistant="""No"""

user="""A: "(*@\textcolor{blue}{\{source\}}@*)"; B: "(*@\textcolor{blue}{\{translation\}}@*)""""

\end{codelisting}

\begin{codelisting} {\tplPlui{Rev} new template for instruction following}     
user= """You will receive two paragraphs P1 and P2 and an instruction I. P1 is a paragraph written for a scientific article, I is an instruction on how to revise P1 and P2 is a revised version of P1. Did P2 address the instruction I? Answer with only one word "Yes" or "No"."""

assistant= """Please provide the data for me to evaluate."""

(*@\textcolor{orange}{<Gold reference can be added here as one-shot>}@*)
user="""P1: "(*@\textcolor{blue}{\{original\}}@*)"; I: "(*@\textcolor{blue}{\{instruction\}}@*)"; P2: "(*@\textcolor{blue}{\{gold\}}@*)""""

assistant= """Yes"""
(*@\textcolor{orange}{<end of gold reference inclusion>}@*)

user="""P1: "(*@\textcolor{blue}{\{original\}}@*)"; I: "(*@\textcolor{blue}{\{instruction\}}@*)"; P2: "(*@\textcolor{blue}{\{hypothese\}}@*)""""

\end{codelisting}

%% file: tables/Nile_results_Mis-Ref.tex
\begin{table}[H]
  \centering
    
\begin{tabular}{l|c|cccc}
\hline
\textbf{model}&\textbf{syst}&\textbf{thr.}&\textbf{f1}&\textbf{prec.}&\textbf{rec.}\\
\midrule
levenshtein         & source &0.17&0.72&0.63&0.84\\
bleu                & source &0.0&0.7&0.54&\textbf{1.0}\\
meteor              & source &0.02&0.74&0.65&0.85\\
bertscore           & source &0.43&0.73&0.66&0.82\\
modern\_bertscore   & source &0.65&0.7&0.54&\textbf{1.0}\\
\midrule
\tplPlui{Net}        & source &-1.91&0\textbf{.78}&0.67&0.93\\
\tplPlui{Net}        & source &0&0.77&\textbf{0.71}&0.84\\
\midrule
\midrule
levenshtein         &Mistral&0.13&0.95&0.91&0.99\\ 
bleu                &Mistral&0.0&0.95&0.9&1.0\\
meteor              &Mistral&0.0&0.95&0.9&1.0\\
bertscore           &Mistral&0.38&0.95&0.9&1.0\\
modern\_bertscore   &Mistral&0.65&0.95&0.9&1.0\\
\midrule
\tplPlui{Net}       &Mistral&-14.74&0.95&0.9&1.0\\
\tplPlui{Net}        &Mistral&0&0.92&\textbf{0.95}&0.89\\
\bottomrule

\end{tabular}
  \caption{Metrics evaluation for the Mistral~\citep{jiangMistral7B2023} generated translations and the English source intents. \phiquatreDef{} is the perplexity model used in \tplPlui{Net}.}
  \label{tab:nile-mis-ref-results_Mistal_Reference}
\end{table}


%% file: appendix/distrib_threshold.tex
\begin{figure}[H]
\centering
    \begin{subfigure}{.80\linewidth}
        \includegraphics[trim={0 0 0 0.85cm},clip,width=1\linewidth]{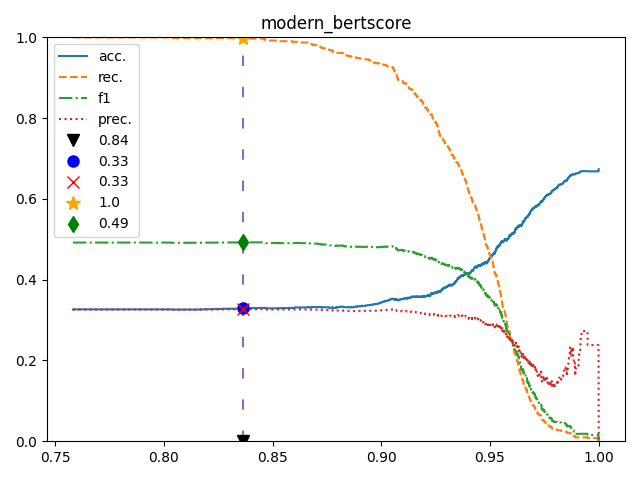}
        \caption{}
        \label{fig:distrib_tytgat_a}
    \end{subfigure}
    \begin{subfigure}{.80\linewidth}
        \includegraphics[trim={0 0 0 0.85cm},clip,width=1\linewidth]{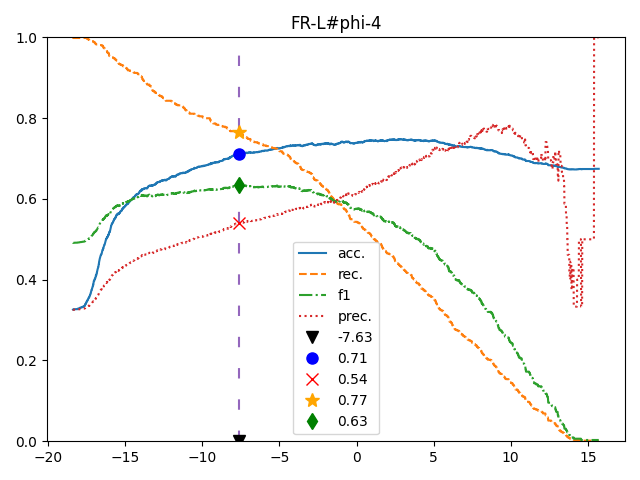}
        \caption{}
        \label{fig:distrib_tytgat_c}
        
    \end{subfigure}
    \caption{
    Score distribution of \ModBScore{} (a) and \tplPlui{Fr} (b).
    The \textcolor{blue}{blue}, \textcolor{orange}{orange}, \textcolor{red}{red} and \textcolor{green}{green} curves denote respectively the \textcolor{blue}{accuracy}, \textcolor{orange}{recall}, \textcolor{red}{precision} and \textcolor{green}{F1-score} according to the decision threshold.
    Emphasis is placed on the maximum F1-score obtained by the metric.
    }
    \label{fig:distrib_tytgat}
\end{figure}

To better understand the results from Section~\ref{sec:res_classif}, Figure~\ref{fig:distrib_tytgat} compares the score distributions of \ModBScore{} and \tplPlui{Fr} on the French paraphrase detection task. 
As shown in Figure~\ref{fig:distrib_tytgat_a}, \ModBScore{} assigns predominantly high similarity scores, and its accuracy curve does not decrease. 
The 0.75 on the x-axis is the lowest Modern BertScore across each pair of sentences in the French paraphrase dataset.
The full 0-to-1 version would be flat from 0 to 0.85.
This graph may look a bit disturbing because it does not show the distribution of scores.
The average Modern BertScore of pairs is 0.95 (with a 95\% confidence interval of 0.00).
Most of the non-paraphrase pairs have a score greater than paraphrase ones and the pair with the higher score is labelled as non-paraphrase.
This indicates that it fails to assign higher similarity scores to sentence pairs that convey the same meaning.
In contrast, \tplPlui{Fr} (Figure~\ref{fig:distrib_tytgat_c}) successfully differentiates paraphrases and non-paraphrases, assigning negative scores to the latter. 
However, the rapid decrease in recall reveals that the model also produces a number of false negative pairs labelled as paraphrases but judged as non-paraphrases, indicating room for improvement.

\begin{figure}[H]
\centering
    \begin{subfigure}{.8\linewidth}
        \includegraphics[trim={0 0 0 13.6mm},clip,width=\linewidth]{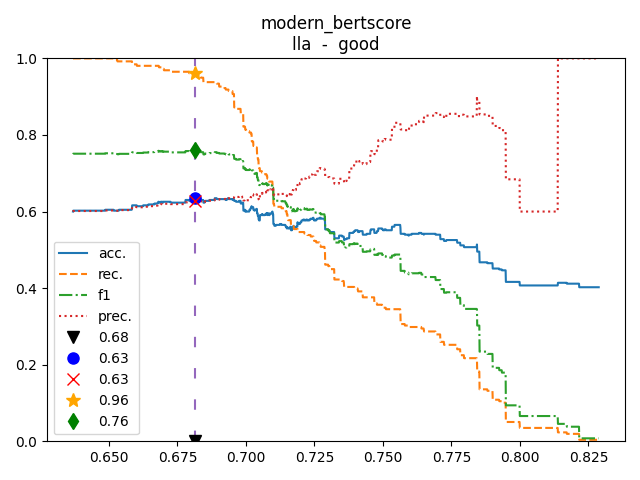} 
        \subcaption{}
        \label{fig:modbert-something}
    \end{subfigure}
    \begin{subfigure}{.8\linewidth}
        \includegraphics[trim={0 0 0 13.6mm},clip,width=\linewidth]{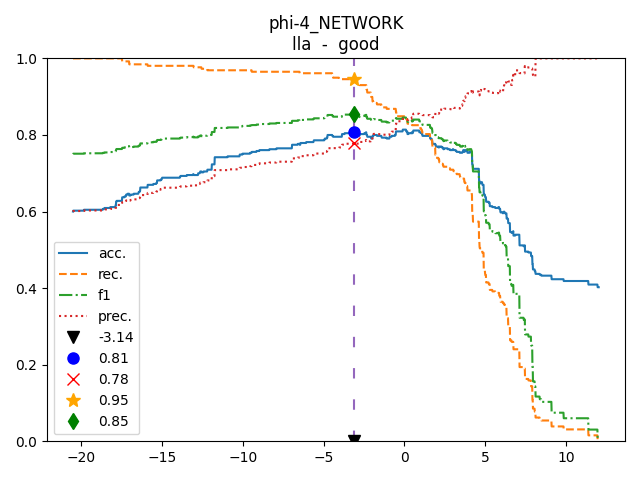}
        \subcaption{}
        \label{fig:parapluie-something}
    \end{subfigure}

\caption{
    \textcolor{blue}{Accuracy}, \textcolor{orange}{recall}, \textcolor{red}{precision} and  \textcolor{green}{F1-score} distribution over different threshold values for \ModBScore\ (a) and  \tplPlui{Net} (b).
    Emphasis is placed on the maximum F1-score obtained by the metric.
}
\label{fig:classification-nile}
\end{figure}

Figure~\ref{fig:classification-nile} presents precision, recall, accuracy, and F1-score across different threshold values for \ModBScore{} and \tplPlui{Net}. 

For \tplPlui{Net} (Figure~\ref{fig:parapluie-something}), the precision increases with the threshold, reducing false positives.
Accuracy and F1 initially increase for threshold values between approximately $-20$ and $0$, then decline rapidly beyond $0$. 
In contrast, \ModBScore{} (Figure~\ref{fig:modbert-something}) shows a slower and shorter initial rise in accuracy and F1, followed by a continuous decrease.
Figures~\ref{fig:apx:levenshtein}, \ref{fig:apx:meteor}, \ref{fig:apx:bertscore}, \ref{fig:apx:bleu}, \ref{fig:apx:modern_bertscore},
\ref{fig:apx:phi-4_FS-DIRECT}, \ref{fig:apx:phi-4_NETWORK} show the different classification results with all the combinations of metrics.

\input{figures/classification-with-different-models}

%% file: figures/classification-with-different-models.tex
\begin{figure}[H]
    \centering
    \begin{minipage}[b]{0.70\textwidth}
        \centering
        \includegraphics[trim={0 0 0 1.36cm},clip,width=1\linewidth]{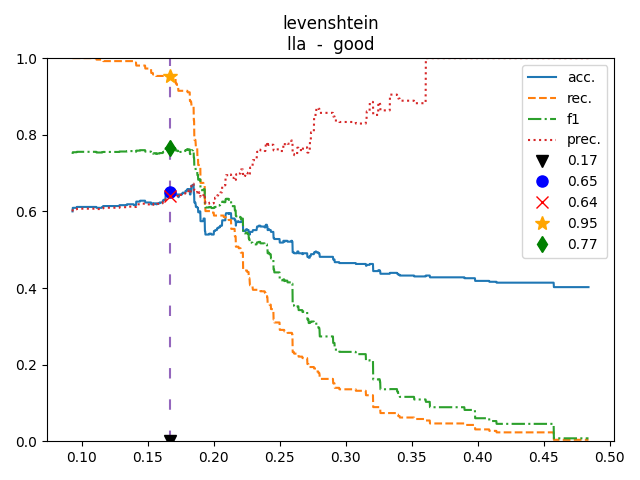}
        \subcaption{Llama translations}
    \end{minipage}%
    \hfill
    \begin{minipage}[b]{0.70\textwidth}
        \centering
        \includegraphics[trim={0 0 0 1.36cm},clip,width=1\linewidth]{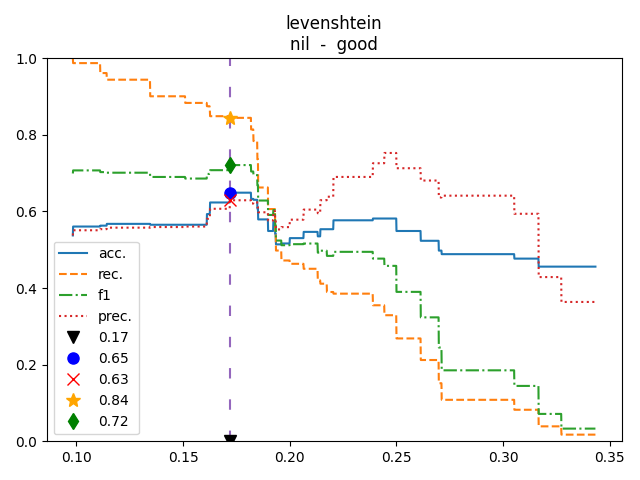}
        \subcaption{English source intents}
    \end{minipage}%
    \hfill
    \begin{minipage}[b]{0.70\textwidth}
        \centering
        \includegraphics[trim={0 0 0 1.36cm},clip,width=1\linewidth]{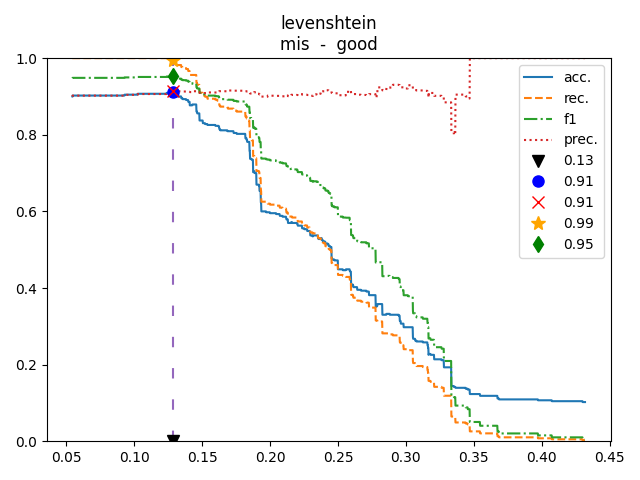}
        \subcaption{Mistral translations}
    \end{minipage}%
    \hfill
\caption{Classification with \Lev{}.}
\label{fig:apx:levenshtein}
\end{figure}

\begin{figure}[H]
    \centering
    \begin{minipage}[b]{0.70\textwidth}
        \centering
        \includegraphics[trim={0 0 0 1.36cm},clip,width=1\linewidth]{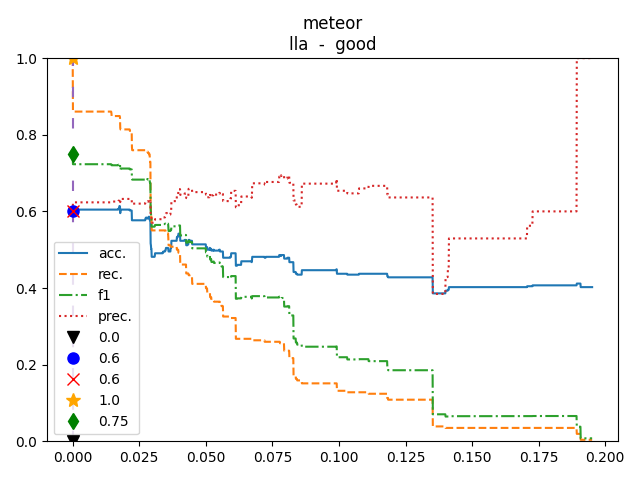}
        \subcaption{Llama translations}
    \end{minipage}%
    \hfill
    \begin{minipage}[b]{0.70\textwidth}
        \centering
        \includegraphics[trim={0 0 0 1.36cm},clip,width=1\linewidth]{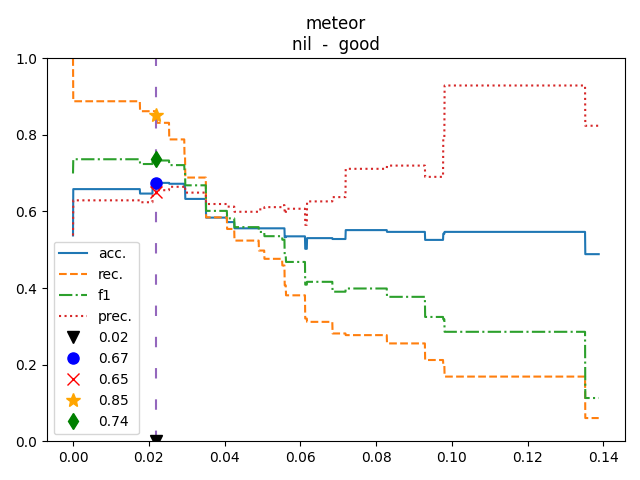}
        \subcaption{English source intents}
    \end{minipage}%
    \hfill
    \begin{minipage}[b]{0.70\textwidth}
        \centering
        \includegraphics[trim={0 0 0 1.36cm},clip,width=1\linewidth]{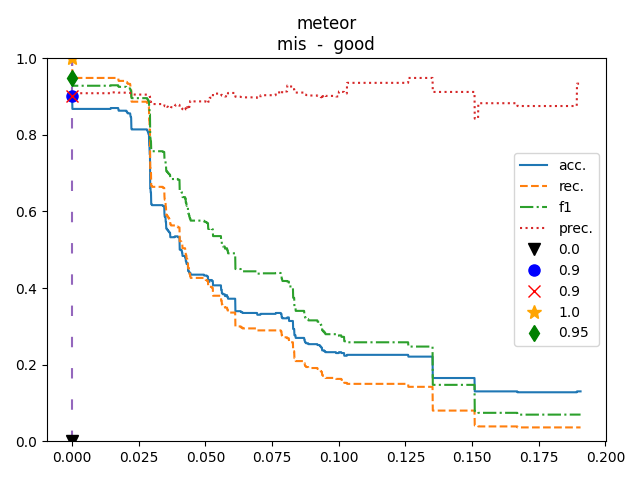}
        \subcaption{Mistral translations}
    \end{minipage}%
    \hfill
\caption{Classification with \Meteor{}.}
\label{fig:apx:meteor}
\end{figure}

\begin{figure}[H]
    \centering
    \begin{minipage}[b]{0.70\textwidth}
        \centering
        \includegraphics[trim={0 0 0 1.36cm},clip,width=1\linewidth]{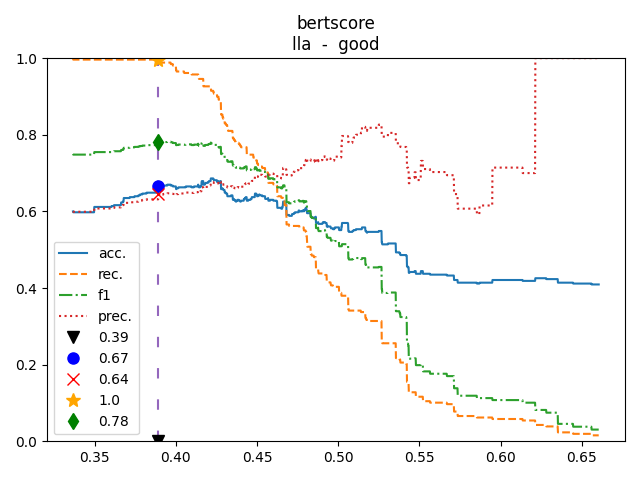}
        \subcaption{Llama translations}
    \end{minipage}%
    \hfill
    \begin{minipage}[b]{0.70\textwidth}
        \centering
        \includegraphics[trim={0 0 0 1.36cm},clip,width=1\linewidth]{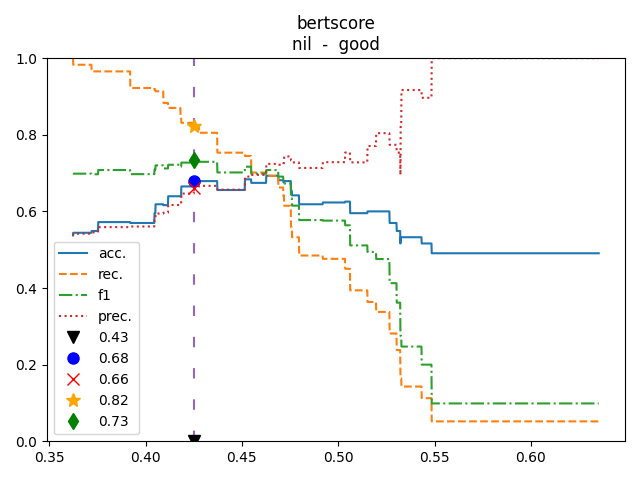}
        \subcaption{English source intents}
    \end{minipage}%
    \hfill
    \begin{minipage}[b]{0.70\textwidth}
        \centering
        \includegraphics[trim={0 0 0 1.36cm},clip,width=1\linewidth]{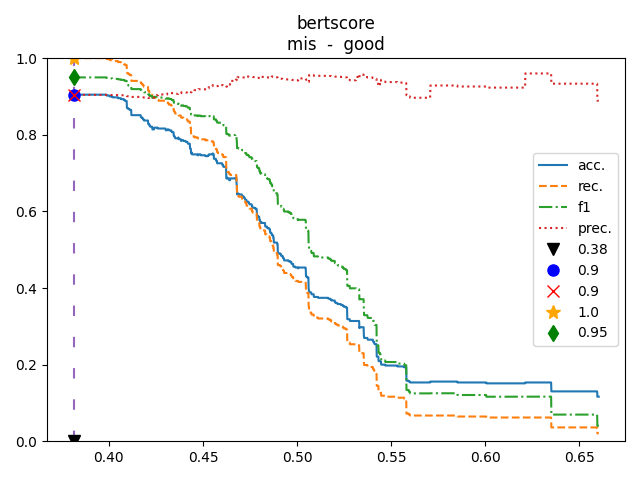}
        \subcaption{Mistral translations}
    \end{minipage}%
    \hfill
\caption{Classification with \BertScore{}.}
\label{fig:apx:bertscore}
\end{figure}

\begin{figure}[H]
    \centering
    \begin{minipage}[b]{0.70\textwidth}
        \centering
        \includegraphics[trim={0 0 0 1.36cm},clip,width=1\linewidth]{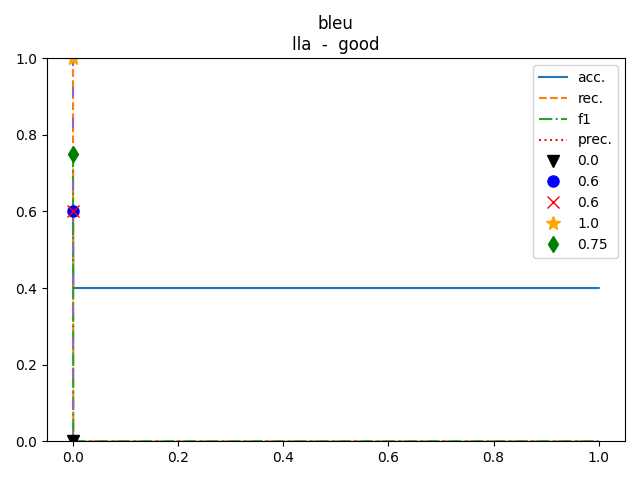}
        \subcaption{Llama translations}
    \end{minipage}%
    \hfill
    \begin{minipage}[b]{0.70\textwidth}
        \centering
        \includegraphics[trim={0 0 0 1.36cm},clip,width=1\linewidth]{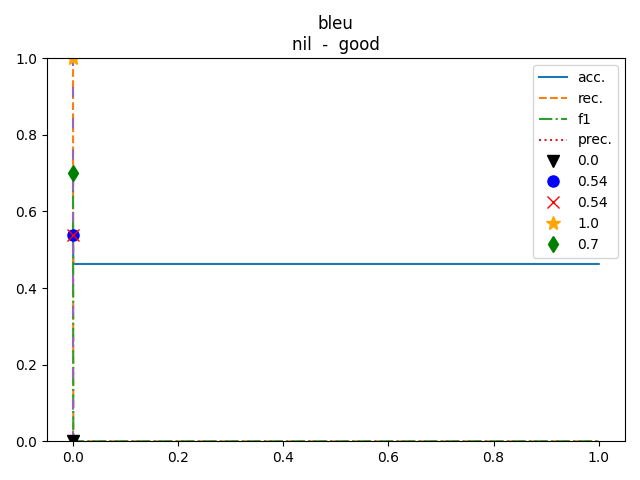}
        \subcaption{English source intents}
    \end{minipage}%
    \hfill
    \begin{minipage}[b]{0.70\textwidth}
        \centering
        \includegraphics[trim={0 0 0 1.36cm},clip,width=1\linewidth]{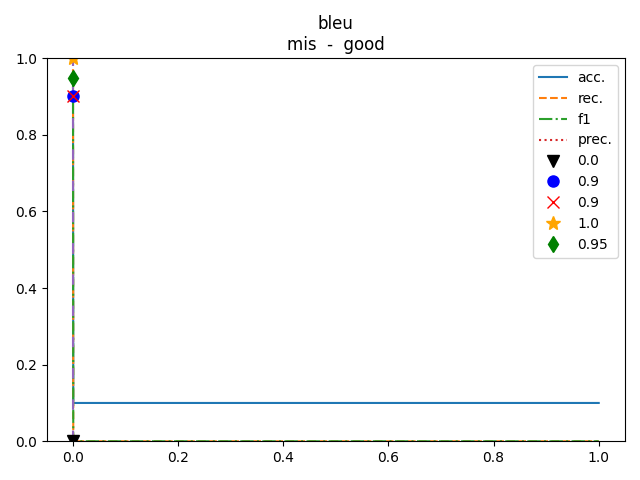}
        \subcaption{Mistral translations}
    \end{minipage}%
    \hfill
\caption{Classification with \Bleu{}.}
\label{fig:apx:bleu}
\end{figure}

\begin{figure}[H]
    \centering
    \begin{minipage}[b]{0.70\textwidth}
        \centering
        \includegraphics[trim={0 0 0 1.36cm},clip,width=1\linewidth]{figures/nile/llama/modern_bertscore_good.png}
        \subcaption{Llama translations}
    \end{minipage}%
    \hfill
    \begin{minipage}[b]{0.70\textwidth}
        \centering
        \includegraphics[trim={0 0 0 1.36cm},clip,width=1\linewidth]{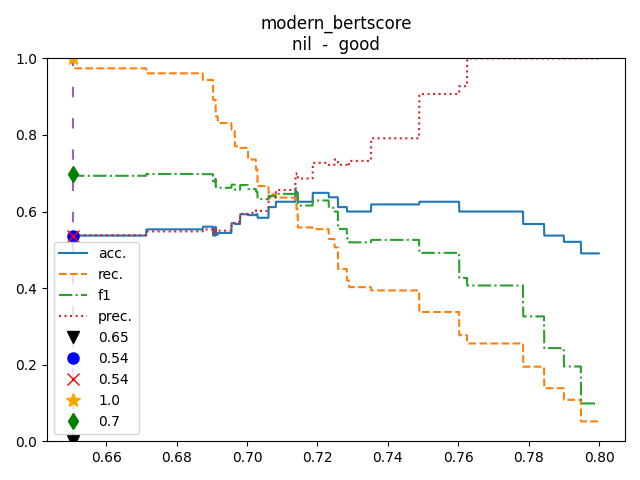}
        \subcaption{English source intents}
    \end{minipage}%
    \hfill
    \begin{minipage}[b]{0.70\textwidth}
        \centering
        \includegraphics[trim={0 0 0 1.36cm},clip,width=1\linewidth]{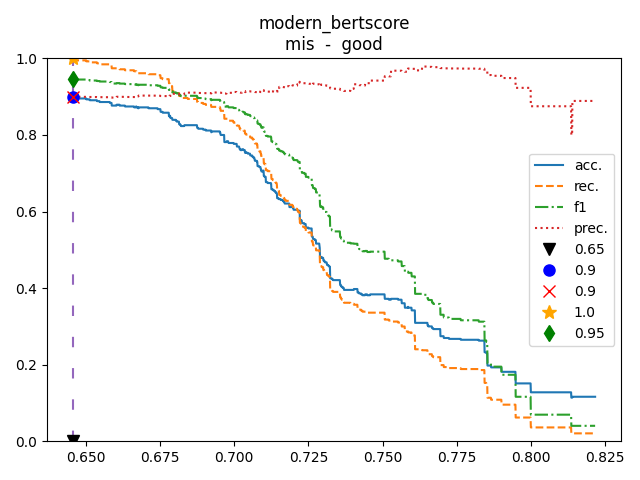}
        \subcaption{Mistral translations}
    \end{minipage}%
    \hfill
\caption{Classification with \ModBScore{}.}
\label{fig:apx:modern_bertscore}
\end{figure}

\begin{figure}[H]
    \centering
    \begin{minipage}[b]{0.70\textwidth}
        \centering
        \includegraphics[trim={0 0 0 1.36cm},clip,width=1\linewidth]{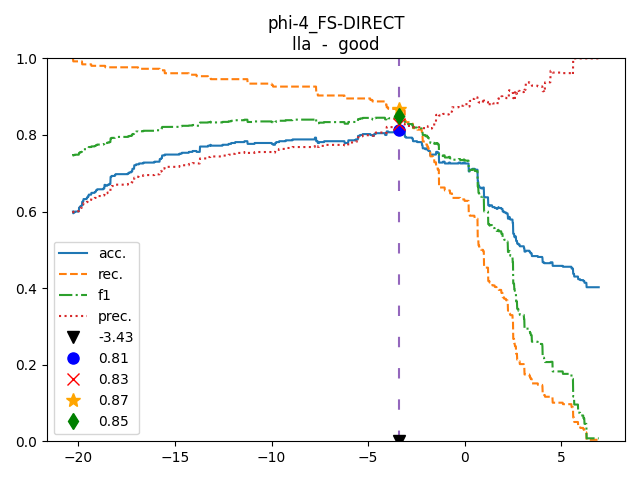}
        \subcaption{Llama translations}
    \end{minipage}%
    \hfill
    \begin{minipage}[b]{0.70\textwidth}
        \centering
        \includegraphics[trim={0 0 0 1.36cm},clip,width=1\linewidth]{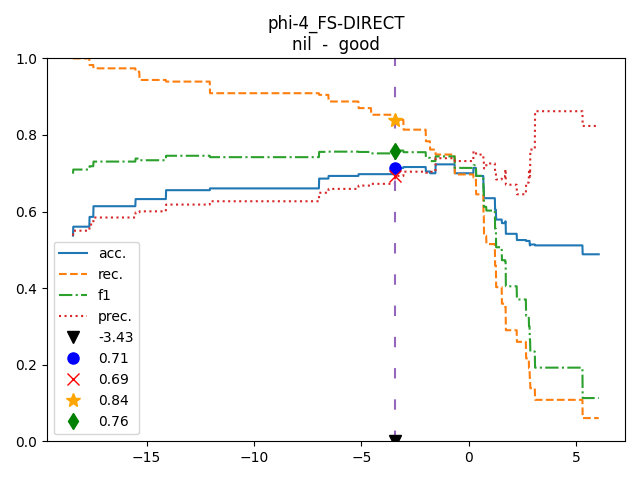}
        \subcaption{English source intents}
    \end{minipage}%
    \hfill
    \begin{minipage}[b]{0.70\textwidth}
        \centering
        \includegraphics[trim={0 0 0 1.36cm},clip,width=1\linewidth]{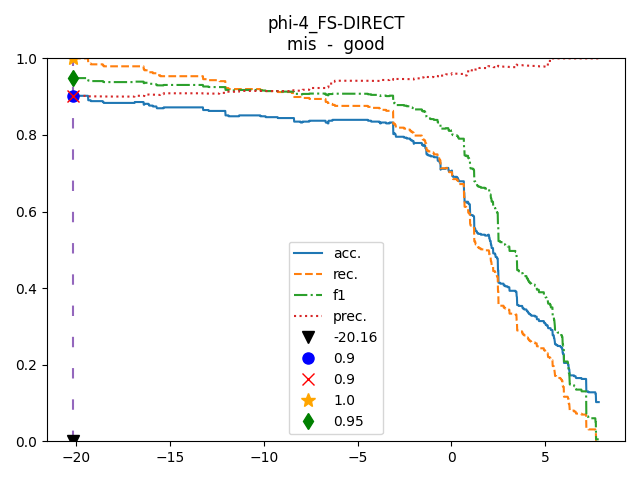}
        \subcaption{Mistral translations}
    \end{minipage}%
    \hfill
\caption{Classification with metric \tplPlui{Para} with \phiquatreDef{}.}
\label{fig:apx:phi-4_FS-DIRECT}
\end{figure}

\begin{figure}[H]
    \centering
    \begin{minipage}[b]{0.70\textwidth}
        \centering
        \includegraphics[trim={0 0 0 1.36cm},clip,width=1\linewidth]{figures/nile/llama/phi-4_NETWORK_good.png}
        \subcaption{Llama translations}
    \end{minipage}%
    \hfill
    \begin{minipage}[b]{0.70\textwidth}
        \centering
        \includegraphics[trim={0 0 0 1.36cm},clip,width=1\linewidth]{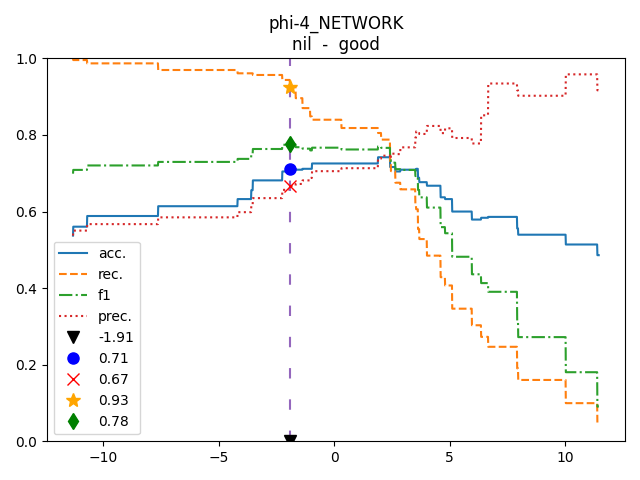}
        \subcaption{English source intents}
    \end{minipage}%
    \hfill
    \begin{minipage}[b]{0.70\textwidth}
        \centering
        \includegraphics[trim={0 0 0 1.36cm},clip,width=1\linewidth]{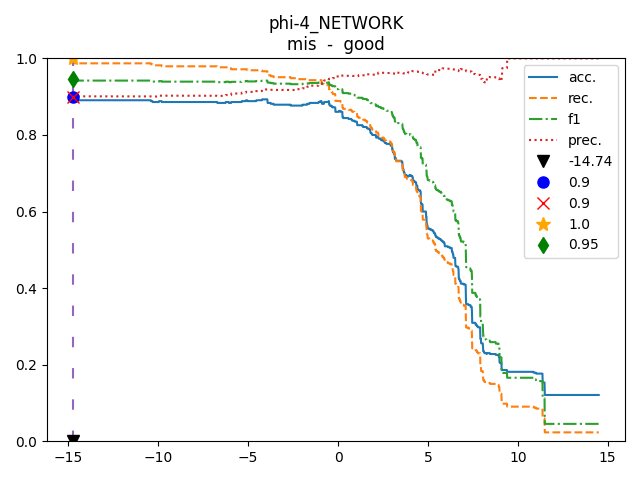}
        \subcaption{Mistral translations}
    \end{minipage}%
    \hfill
\caption{Classification with metric \tplPlui{Net} with \phiquatreDef{}.}
\label{fig:apx:phi-4_NETWORK}
\end{figure}

%% file: tables/appendix_align_rev.tex
\begin{table}[H]
  \centering

\begin{tabular}{ c l| cc | cc | cc}
    \toprule
     \multicolumn{2}{c|}{\multirow{2}{*}{\textbf{Judge}}} & \multicolumn{2}{c|}{\textbf{Pair acc.}} & \multicolumn{2}{c|}{{\large \textbf{$V$}}} & \multicolumn{2}{c}{{\Large \textbf{$\kappa$}}} \\
               &    &   & w.g. &   & w.g. &   & w.g. \\
    \midrule
     \parbox[t]{2mm}{\multirow{4}{*}{\rotatebox[origin=c]{90}{{\scriptsize \textbf{\tplPlui{Rev}}} }}}
  & \LlamaSDef & \textbf{0.61} & \textbf{0.62} & \textbf{0.31} & \textbf{0.32} & \textbf{0.32} & \textbf{0.34}\\
  & \phiquatreDef & \textbf{0.61} & \underline{0.61} & \textbf{0.31} & \textbf{0.32} & \textbf{0.32}  & \underline{0.33}\\
  & \LlamaH    & 0.58 & 0.59 & 0.27 & 0.29 & 0.27 & 0.29\\
  & \mistral   & 0.55 & 0.59 & 0.24 & 0.28 & 0.22 & 0.28\\
     
     \midrule

\parbox[t]{2mm}{\multirow{4}{*}{\rotatebox[origin=c]{90}{{\scriptsize \textbf{\tplPlui{Para}}} }}}
  & \LlamaSDef    & 0.52 &   & 0.20 &   & 0.17 & \\
  & \phiquatreDef & 0.52 &   & 0.21 &   & 0.15 &  \\
  & \LlamaH    & 0.52 &   & 0.21 &   & 0.17 &  \\
  & \mistral   & 0.54 &   & 0.22 &   & 0.20 &   \\

     \midrule
     
     \parbox[t]{2mm}{\multirow{5}{*}{\rotatebox[origin=c]{90}{{\scriptsize \textbf{LLM choice} }}}}
     & \gptIV    & \underline{0.59} & 0.60 & 0.28 & \underline{0.30} & \underline{0.30} & \underline{0.33}\\
  & \gptIVmini & 0.57 & 0.58 & 0.25 & 0.26 & 0.27 & 0.29\\
  & \LlamaSDef & \underline{0.59} & 0.60 & 0.28 & \underline{0.30} & \underline{0.30} & 0.31\\
  & \phiquatreDef & 0.53 & 0.55 & 0.25 & 0.27 & 0.24 & 0.27\\
  & \LlamaH    & 0.54 & 0.51 & 0.21 & 0.18 & 0.20 & 0.17\\
  & \mistral   & 0.53 & 0.53 & 0.20 & 0.17 & 0.17 & 0.16\\
  
  \midrule
  
  \parbox[t]{2mm}{\multirow{5}{*}{\rotatebox[origin=c]{90}{{\scriptsize \textbf{LLM likert}}}}}
   &  \gptIV   & 0.54 & 0.54 & 0.28 & 0.27 & 0.28 & 0.27\\
  & \gptIVmini & 0.45 & 0.51 & 0.28 & 0.27 & 0.21 & 0.23\\
  & \LlamaSDef & 0.44 & 0.50 & \underline{0.29} & 0.27 & 0.19 & 0.23\\
  & \phiquatreDef & 0.45 & 0.52 & 0.30 & 0.29 & 0.21 & 0.26\\
  & \LlamaH    & 0.43 & 0.45 & 0.18 & 0.18 & 0.15 & 0.15\\
  & \mistral   & 0.33 & 0.28 & 0.16 & 0.10 & 0.09 & 0.05\\
    \bottomrule
\end{tabular}

  \caption{Alignment of LLM-based metrics with human judgements.
Pairwise accuracy and Cramér’s V are defined on [0:1] and Cohen’s Kappa on [-1:1]. w.g. (with gold) indicates that the reference revision is provided. For \tplPlui{Para}, the column without gold corresponds to the \tplPlui{Para} scores between the original and generated paragraphs.
Perplexity models used are Llama-3, Phi-4, Mistral and GPT-4o \citep{grattafioriLlama3Herd2024, abdinPhi4TechnicalReport2024, jiangMistral7B2023, openai2024gpt4ocard}.
}
  \label{tab:rev_all_results_appendix}
\end{table}

%% file: appendix/netpluie.tex
\citet{MUNSON2025} introduced \ac{NEAT}, a methodology used to create a large-scale corpus of aligned English–\ac{Nile} intents.
In their methodology, the authors formalise the $align(i, n)$ function which assesses if a Nile intent $(i)$ and a natural language statement $(n)$ share the same underlying meaning.
So far, $align(\cdot, \cdot)$ has only been reliably implemented through manual expert annotation, which limits scalability and reproducibility.
To validate \tplPlui{Net} as a semantic alignment function, we require two conditions to be verified, presented in Equations \ref{equ:implication-aa} and \ref{equ:implication-app}.

\begin{equation}
    align(i,n_1) \land align(i,n_2) \implies \texttt{\tplPlui{*}}(n_1, n_2)>T \land \texttt{\tplPlui{*}}(n_2, n_1)>T
\label{equ:implication-aa}
\end{equation}

\begin{equation}
    align(i,n_1) \land \texttt{\tplPlui{*}}(n_1, n_2) > T \land \texttt{\tplPlui{*}}(n_2, n_1)>T \implies align(i,n_2)
\label{equ:implication-app}
\end{equation}

Therefore, if two sentences ($n_1$ and $n_2$) share the same meaning and one of them is aligned with an intent, the second one should also be aligned with this intent. 
This is made possible because the intent language is unambiguous; indeed, this relation is not necessarily true with natural language.
We exploit from the non-symmetrical property of \tplPlui{*} and compute the score \tplPlui{Net}($n_1$,$n_2$) and \tplPlui{Net}($n_2$,$n_1$).
However, when used as the alignment measure between natural language and formal intent language, we only use it once with the formal intent as the reference. 
This is because natural language expressions usually capture more details than formal languages; these additional details captured in the expression should still be considered as part of a good translation. 
However, the metric penalises the suppression of information between two expressions, meaning that the loss of information that is not relevant to the actual network configurations when going from natural language to Nile is penalised by the metric.
We consider the Mistral and Llama generated translations as our two sources of natural language expressions and count how many times our two implications are not respected.
Overall, of the 430 Nile-NL triplets, 56 do not verify condition \ref{equ:implication-aa} and 7 fail condition \ref{equ:implication-app}. 
Examples of the translations that failed the tests can be found in Table~\ref{tab:bad_logic-aa} and Table~\ref{tab:bad_logic-app}.

Despite only 15\% of the triplets failing our imposed conditions, this makes it difficult to use this metric comfortably as a semantic alignment function without further verification on the generated translations before acceptance.
Fortunately, the most critical condition is condition \ref{equ:implication-app}, which enables the deduction of an alignment given an existing alignment and a paraphrased sentence, and this condition is satisfied in 98\% of the cases.
So we can consider \tplPlui{Net} as the closest solution we have to an accurate automatic semantic alignment function between natural language and a formal intent language, making it the best solution available.

\newpage

\input{tables/bad_translation}

%% file: tables/bad_translation.tex
\begin{table}[H]
  \centering
  \small
\begin{tabular}{p{0.3\linewidth}|p{0.3\linewidth}|p{0.3\linewidth}}
\hline
Nile sentence&Translation with Llama~\citep{grattafioriLlama3Herd2024}&Translation with Mistral~\citep{jiangMistral7B2023}\\ \hline
\hline
for endpoint('dorms'), endpoint('wireless') set quota('any', '6', 'gb/d') start hour('00:00') end hour('23:59')&The ResNet and NYURoam wireless networks have a 6 GB per day quota&The ResNet and wireless network have a daily cap of 6 GB per day, from midnight to midnight\\ \hline
for endpoint('dorms'), endpoint('wireless') set quota('any', '6', 'gb/d') start hour('00:00') end hour('23:59')&The ResNet cap is currently 6 GB per day&ResNet and UIC wireless have a daily cap of 6 GB for all incoming and outgoing traffic\\ \hline
for endpoint('dorms'), endpoint('wireless') set quota('any', '6', 'gb/d') start hour('00:00') end hour('23:59')&The ResNet cap is currently 6 GB per day&ResNet and wireless users have a daily quota of 6 GB, with no time restrictions\\ \hline
for endpoint('dorms'), endpoint('wireless') set quota('any', '6', 'gb/d') start hour('00:00') end hour('23:59')&The ResNet cap is currently 6 gigabytes per day&The ResNet cap is currently 6 gigabytes per day for each 24 hour period of combined incoming and outgoing traffic\\ \hline
for endpoint('guests') block service('file transfer')&Guests are not allowed to run file transfer services&Guests are not allowed to transfer files\\ \hline
for endpoint('guests') block service('file transfer')&Guests are not allowed to use file transfer protocols&Guests are not allowed to transfer files\\ \hline
for endpoint('host') set bandwidth('max', '3', 'mbps')&The individual will have a 3 Mbps bandwidth rate limit&The host will have a 3 Mbps bandwidth rate limit\\ \hline
for endpoint('host') set bandwidth('max', '4', 'mbps')&Hosts' network is rate-limited to 4 Mb/s per user&Each host is limited to 4 Mb/s\\ \hline
for endpoint('host') set bandwidth('max', '4', 'mbps')&Hosts' network is rate-limited to 4 Mb/s per user&Hosts are rate-limited to 4 Mb/s\\ \hline
for endpoint('host') set bandwidth('max', '4', 'mbps')&Internet is limited to 4 Mbps per host&Each host is limited to 4 Mbps\\ \hline
for endpoint('network') add middlebox('network border system'), middlebox('ips'), middlebox('firewall'), middlebox('unit firewall') &The network is protected by a firewall, a network border system, and a unit firewall&The network has multiple layers of security, including a network border system, intrusion prevention system, firewall, and unit firewall\\ \hline
for endpoint('network') add middlebox('overuse notification') set quota('any', '10', 'gb/wk')&The network will send an email to the user when they have used 10 GB of bandwidth in a week&If you exceed 10 GB of network usage in a week, you will receive a notification\\ \hline
for endpoint('network') block service('NETBIOS')&The UVA More Secure Network does not allow NetBIOS&NETBIOS is blocked by the University of Virginia's firewall\\ \hline
for endpoint('network') block service('NETBIOS')&The network firewall does prevent some applications from working in their normal manner, for example, NetBIOS&NETBIOS is blocked by the network firewall\\ \hline
for endpoint('network') block service('NETBIOS')&The network firewall does prevent some applications from working in their normal manner, for example, NetBIOS&The network firewall blocks the use of the NETBIOS protocol\\ \hline
for endpoint('network') block traffic('H323 video conferencing')&H323 video conferencing is blocked by the University network&H323 video conferencing is blocked by the network\\ \hline
for endpoint('network') block traffic('H323 video conferencing')&H323 video conferencing is blocked&H323 video conferencing is blocked on the network\\ \hline
for endpoint('network') set quota('any', '10', 'gb/d')&The University of Illinois at Urbana-Champaign has a 10 GB per day limit on all combined traffic&The network has a 10 GB per day data transfer limit\\ \hline
\end{tabular}

  \caption{A few examples of bad translations that do not respect implication \ref{equ:implication-aa}, both translations are aligned to the same intent expression, but \PPluie{} does not consider the translations to be paraphrases.}
  \label{tab:bad_logic-aa}
\end{table}

\begin{table}[H]
  \centering
  \small
    
\begin{tabular}{p{0.3\linewidth}|p{0.3\linewidth}|p{0.3\linewidth}}
\hline
Nile sentence&Translation with Llama~\citep{grattafioriLlama3Herd2024}&Translation with Mistral~\citep{jiangMistral7B2023}\\ \hline
\hline
for endpoint('dorms') block protocol('nat')&NAT is not allowed on the ResNet&NAT is blocked on the ResNet\\ \hline
for endpoint('dorms'), endpoint('wireless') block protocol('ftp'), protocol('telnet'), traffic('peer2peer'), protocol('http'), service('irc'), protocol('dns'), service('email')&You may not run any services (e.g., FTP services, telnet, P2P file sharing servers, publishing files via generic file services, etc.) via NYU's ResNet or NYURoam wireless networks&ResNet and wireless networks are blocked from using FTP, Telnet, P2P, HTTP, IRC, DNS, and email services\\ \hline
for endpoint('dorms'), endpoint('wireless') block protocol('ftp'), service('listserver'), traffic('peer2peer')&The University of Washington does not allow FTP, Listserv, or Peer-to-Peer traffic on the ResNet or UW Wireless&ResNet and UIC wireless do not allow FTP, listserv, or peer-to-peer traffic\\ \hline
for endpoint('dorms'), endpoint('wireless') block protocol('ftp'), service('listserver'), traffic('peer2peer')&You may not run FTP, listserver, peer-to-peer, or file sharing/downloading program&ResNet and wireless users cannot use FTP, listserv, or peer-to-peer file sharing\\ \hline \hline
for endpoint('dorms'), endpoint('wireless') set quota('any', '6', 'gb/d') start hour('00:00') end hour('23:59')&Dorms and wireless have a quota of 6 GB per day&The quota for dorms and wireless is 6 GB per day, with no time restrictions\\ \hline
for endpoint('dorms'), endpoint('wireless') set quota('any', '6', 'gb/d') start hour('00:00') end hour('23:59')&Students have a quota of 6 GB per day The quota is reset at midnight&ResNet and wireless users have a daily quota of 6 GB, with no time restrictions\\ \hline
for endpoint('dorms'), endpoint('wireless') set quota('any', '6', 'gb/d') start hour('00:00') end hour('23:59')&The ResNet and NYURoam wireless networks have a 6 GB per day quota&ResNet and wireless network have a daily quota of 6 GB, available 24/7\\ \hline
\end{tabular}

  \caption{Bad translations that do not respect implication \ref{equ:implication-app}: One translation is aligned to the Nile expression, both translations are paraphrases, but, on lines 1 to 4, \PPluie{} does not consider Llama's translation to be aligned to the Nile expression, and on lines 5 to 7, \PPluie{} does not consider Mistral's translation to be aligned to the Nile expression.}
  \label{tab:bad_logic-app}
\end{table}

%% file: tables/add_parap.tex
\begin{table}[H]
  \centering
\begin{tabular}{l|cccccc}
    \toprule
     & \multicolumn{6}{c}{French Paraphrase Detection} \\
    \textbf{Metric}&\textbf{Acc.}&\textbf{Rec.}&\textbf{Prec.}&\textbf{F1}&\textbf{GPUs}&\textbf{Runtime} \\ 

   \midrule
    LLM-Yes/No \phiquatreDef~ (French)   & 0.74 & 0.53 & 0.61 & 0.57 & MI300 x1 & 27 min \\
    LLM-Yes/No \LlamaSDef~ (French)      & 0.71 & 0.57 & 0.55 & 0.56 & MI300 x2 & 57 min\\
    \bottomrule
\end{tabular}
  \caption{Results of LLM-as-a-judge approaches with a French prompt.}
  \label{tab:llmjudge_para_appendix}
\end{table}

%% file: appendix/additionnal_paraphrase.tex
\begin{table}[H]
  \centering
\begin{tabular}{l|ccccc}
    \toprule
     & \multicolumn{5}{c}{French Paraphrase Classification} \\
    \textbf{Metric}&\textbf{Thr.}&\textbf{Acc.}&\textbf{Rec.}&\textbf{Prec.}&\textbf{F1} \\ 

   \midrule
    \tplPlui{Para} SmolLM2-135M-Instruct & 0 & 0.67 & N/A & 0.00 & 0.00 \\
    \tplPlui{Para} SmolLM2-135M-Instruct & -2.31 & 0.33 & 0.33 & 1.00 & 0.49 \\
    \midrule
    \tplPlui{Para} SmolLM2-360M-Instruct & 0 & 0.67 & 0.35 & 0.03 & 0.06 \\
    \tplPlui{Para} SmolLM2-360M-Instruct & -1.26 & 0.33 & 0.33 & 1.00 & 0.49 \\
    \midrule
    \tplPlui{Para} SmolLM2-1.7B-Instruct & 0 & 0.61 & 0.41 & 0.44 & 0.43 \\
    \tplPlui{Para} SmolLM2-1.7B-Instruct & -0.75 & 0.41 & 0.35 & 0.92 & 0.51 \\
    \midrule
    \tplPlui{Para} internlm2-chat-1\_8b & 0 & 0.39 & 0.34 & 0.93 & 0.50 \\
    \tplPlui{Para} internlm2-chat-1\_8b & 0.23 & 0.43 & 0.35 & 0.89 & 0.50 \\
    \midrule
    \tplPlui{Para} gemma-2-2b-it & 0 & 0.61 & 0.44 & 0.71 & 0.54 \\
    \tplPlui{Para} gemma-2-2b-it & -0.67 & 0.58 & 0.42 & 0.77 & 0.55 \\
    \midrule
    \tplPlui{Para} Phi-4-mini-instruct & 0 & 0.65 & 0.48 & 0.68 & 0.56 \\
    \tplPlui{Para} Phi-4-mini-instruct & -2.52 & 0.58 & 0.43 & 0.85 & 0.57 \\
    \midrule
    \tplPlui{Para} Mistral-7B-Instruct-v0.2 & 0 & 0.67 & 0.49 & 0.48 & 0.49 \\
    \tplPlui{Para} Mistral-7B-Instruct-v0.2 & -10.71 & 0.59 & 0.43 & 0.76 & 0.55 \\
    \midrule
    \tplPlui{Para} Qwen2.5-7B-Instruct & 0 & 0.71 & 0.55 & 0.59 & 0.57 \\
    \tplPlui{Para} Qwen2.5-7B-Instruct & -13.04 & 0.67 & 0.50 & 0.73 & 0.59 \\
    \midrule
    \tplPlui{Para} aya-expanse-8b & 0 & 0.60 & 0.43 & 0.65 & 0.52 \\
    \tplPlui{Para} aya-expanse-8b & -5.7 & 0.53 & 0.40 & 0.87 & 0.54 \\
    \midrule
    \tplPlui{Para} Llama-3.1-8B-Instruct & 0 & 0.72 & 0.58 & 0.48 & 0.52 \\
    \tplPlui{Para} Llama-3.1-8B-Instruct & -6.52 & 0.64 & 0.47 & 0.79 & 0.59 \\
    \midrule
    \tplPlui{Para} gemma-2-9b-it & 0 & 0.70 & 0.54 & 0.55 & 0.54 \\
    \tplPlui{Para} gemma-2-9b-it & -5.83 & 0.63 & 0.46 & 0.78 & 0.58 \\
    \midrule
    \tplPlui{Para} aya-expanse-32b & 0 & 0.65 & 0.48 & 0.63 & 0.54 \\
    \tplPlui{Para} aya-expanse-32b & -2.65 & 0.62 & 0.45 & 0.74 & 0.56 \\
    \midrule
    \tplPlui{Para} QwQ-32B & 0 & 0.70 & 0.53 & 0.64 & 0.58 \\
    \tplPlui{Para} QwQ-32B & -0.13 & 0.70 & 0.53 & 0.67 & 0.59 \\
    \midrule
    \tplPlui{Para} c4ai-command-r-08-2024 & 0 & 0.70 & 0.59 & 0.29 & 0.39 \\
    \tplPlui{Para} c4ai-command-r-08-2024 & -7.26 & 0.62 & 0.45 & 0.79 & 0.57 \\
    \bottomrule
\end{tabular}
  \caption{Results of \tplPlui{Para} approaches with for French paraphrase classification with different perplexity models.
  In the following order, SmolLM2-Instruct~\citep{allalSmolLM2WhenSmol2025}, internlm2-chat~\citep{caiInternLM2TechnicalReport2024}, gemma-2 ~\citep{teamGemma2Improving2024}, Phi-4-mini-instruct~\citep{microsoftPhi4MiniTechnicalReport2025}, Mistral-7B-Instruct~\citep{jiangMistral7B2023},  Qwen2.5-Instruct~\citep{qwenQwen25TechnicalReport2025}, aya-expanse~\citep{dangAyaExpanseCombining2024}, Llama-3.1-Instruct~\citep{grattafioriLlama3Herd2024}, QwQ~\citep{qwq32b} and c4ai-command-r~\citep{vergaReplacingJudgesJuries2024}.}
  \label{tab:additional_para_appendix}
\end{table}